%% file: PaperForReview.tex
\documentclass[runningheads]{llncs}

\usepackage[dvipsnames,table]{xcolor}

\usepackage[final,year=2024,ID=9733]{eccv}
\usepackage{eccvabbrv}

\usepackage{graphicx}
\usepackage{amsmath}
\usepackage{amssymb}
\usepackage{comment}
\usepackage{booktabs}

\usepackage[accsupp]{axessibility}  

\usepackage[pagebackref,breaklinks,colorlinks,citecolor=eccvblue]{hyperref}
\usepackage{orcidlink}

\usepackage[capitalize]{cleveref}
\crefname{section}{Sec.}{Secs.}
\Crefname{section}{Section}{Sections}
\Crefname{table}{Table}{Tables}
\crefname{table}{Tab.}{Tabs.}

\usepackage{bm}

\newcommand{\myVector}[1]{\bm{\mathit{#1}}}
\newcommand{\argmin}{\mathop{\mathrm{argmin}}\limits}

\usepackage[inkscapeformat=png, inkscapearea=page]{svg}

\usepackage{enumitem}

\usepackage{xcolor}

\definecolor{ibm2}{HTML}{56B4E9}
\definecolor{ibm3}{HTML}{000000}
\definecolor{ibm4}{HTML}{E69F00}

\usepackage{pgfplots}
\pgfplotsset{compat=newest}
\pgfplotscreateplotcyclelist{foo bar}{%
ultra thick, every mark/.append style={solid}, mark=*, color=ibm2\\%
densely dotted, every mark/.append style={solid}, mark=*, color=ibm2\\%
ultra thick, every mark/.append style={solid}, mark=square*, color=ibm3\\%
densely dotted, every mark/.append style={solid}, mark=square*, color=ibm3\\%
ultra thick, every mark/.append style={solid}, mark=triangle*, color=ibm4\\%
densely dotted, every mark/.append style={solid}, mark=triangle*, color=ibm4\\%
}

\usepackage{adjustbox}
\usepackage{array}

\newcolumntype{R}[2]{%
    >{\adjustbox{angle=#1,lap=\width-(#2)}\bgroup}%
    l%
    <{\egroup}%
}
\newcommand*\rot{\multicolumn{1}{R{30}{1em}}}

\usepackage{hhline, multirow}

\usepackage{wrapfig}

\begin{document}

\title{RANRAC: Robust Neural Scene Representations via Random Ray Consensus}

\titlerunning{RANRAC}


\author{
Benno Buschmann \inst{1,2} \and
Andreea Dogaru \inst{2} \and
Elmar Eisemann \inst{1} \and
Michael Weinmann \inst{1} \and
Bernhard Egger \inst{2}
}

\authorrunning{B.~Buschmann et al.}


\institute{Delft University of Technology, Netherlands \and
Friedrich-Alexander-University Erlangen-Nürnberg, Germany}


\maketitle


\begin{abstract}
Learning-based scene representations such as neural radiance fields or light field networks, that rely on fitting a scene model to image observations, commonly encounter challenges in the presence of inconsistencies within the images caused by occlusions, inaccurately estimated camera parameters or effects like lens flare.
To address this challenge, we introduce {RANdom RAy Consensus (RANRAC)}, an efficient approach to eliminate the effect of inconsistent data, thereby taking inspiration from classical RANSAC based outlier detection for model fitting.
In contrast to the down-weighting of the effect of outliers based on robust loss formulations, our approach reliably detects and excludes inconsistent perspectives, resulting in clean images without floating artifacts.
For this purpose, we formulate a fuzzy adaption of the RANSAC paradigm, enabling its application to large scale models. We interpret the minimal number of samples to determine the model parameters as a tunable hyperparameter, investigate the generation of hypotheses with data-driven models, and analyse the validation of hypotheses in noisy environments.
We demonstrate the compatibility and potential of our solution for both photo-realistic robust multi-view reconstruction from real-world images based on neural radiance fields and for single-shot reconstruction based on light-field networks.
In particular, the results indicate significant improvements compared to state-of-the-art robust methods for novel-view synthesis on both synthetic and captured scenes with various inconsistencies including occlusions, noisy camera pose estimates, and unfocused perspectives.
The results further indicate significant improvements for single-shot reconstruction from occluded images.

\textbf{Project Page:} \url{https://bennobuschmann.com/ranrac/}


\end{abstract}

\section{Introduction}
\input{figures/teaser}
3D reconstruction is a classical task in computer vision and computer graphics, which has attracted research for decades. It offers numerous applications,  including autonomous systems, entertainment, design, advertisement, cultural heritage, VR/AR experiences or medical scenarios.
In recent years, neural scene representations and rendering techniques~\cite{Xie2021NeuralFI,tewari2022advances}, including light field networks (LFN)~\cite{sitzmann2021lfns} and neural radiance fields (NeRF)~\cite{mildenhall2020nerf} have demonstrated great performance in single-view and multi-view reconstruction tasks.
The key to the success of such techniques is the coupling of differentiable rendering methods with custom neural field parametrizations of scene properties.
However, a common limitation of neural scene reconstruction methods is their sensitivity to inconsistencies among input images induced by occlusions, inaccurately estimated camera parameters or effects like lens flares.
Despite the use of view-dependent radiance representations to address view-dependent appearance changes, these inconsistencies severely impact local density estimation, resulting in a poor generalization to novel views.
%

%

%
To increase the robustness to potential distractors within the training data, Sabour et al.~\cite{sabour2023robustnerf} recently introduced the use of robust losses in the context of training unconditioned NeRF, where distractors in the training data were modeled as outliers of an optimization problem.
However, the adaptation of this approach to conditioned neural fields (e.g., pixelNeRF \cite{yu2021pixelnerf}) is not obvious, as no optimization takes place during inference, and the input data is constrained to only a few views.
Achieving robustness to data inconsistencies is a well-analyzed problem in computer vision, covered not only by the aforementioned robust loss functions~\cite{barron2019general}, but also other strategies, like the random sample consensus (RANSAC) paradigm \cite{fischler1981random}. The latter is widely employed for fitting models to outlier-heavy data.
The underlying idea is to randomly select subsets of the data to form potential models, evaluating these models against the entire dataset, and identifying the subset that best fits the majority of the data, while disregarding outliers.  
Despite being the state-of-the-art solution to many challenges, RANSAC-based schemes are particularly favoured for the fitting of analytical models with a relatively small amount of tuneable parameters. 
In this paper, we direct our attention to achieving robustness against inconsistencies and occlusions in the observations by using a novel combination of neural scene representation and rendering techniques with dedicated outlier removal techniques such as RANSAC~\cite{fischler1981random}.
While downweighting the influence of distractors based on robust losses~\cite{barron2019general,sabour2023robustnerf} can affect clean samples,  representing details, we aim at improving robustness to distractors by only removing the influence of outliers. 
To this extent, we integrate a RANSAC-based scheme to distinguish inliers and outliers in the data and the inlier-based optimization of the neural fields (Fig.~\ref{fig:teaser}); a stochastic scenario  characterized by a large-scale, data-driven model that exceeds RANSAC's classical convergence expectations.
Instead of guaranteeing convergence to a clean sample set based on a minimal number of samples, we aim for a feasible (cleaner) sample set using a tuneable amount of samples. The proposed algorithm exhibits robustness and versatility, accommodating a wide range of neural fields-based reconstruction methods. 
Our method inherits the strengths of RANSAC, such as the ability to handle various classes of outliers without relying on semantics. Yet, it also inherits the need for sufficiently clean samples and the reliance on an iterative scheme. In practice, the first condition is often fulfilled because typically only some of the perspectives are affected by inconsistencies.
We validate our approach using synthetic data, focusing on the task of multi-class single-shot reconstruction with LFNs~\cite{sitzmann2021lfns}, and observe significant quality improvements over the baseline in the presence of occlusions.
Furthermore, we showcase robust photo-realistic reconstructions of 3D objects using unconditioned NeRFs from sequences of real-world images in the presence of distractors. In comparison to RobustNeRF \cite{sabour2023robustnerf}, we use all available clean data, hence improving the reconstruction quality for single-object scenes.
Code and data are available under \emph{(link follows upon acceptance)}.

\noindent Our key contributions are:
\begin{itemize}
\item a general, robust RANSAC-based reconstruction method applicable to a variety of neural-fields and handling diverse inconsistencies
\item an analysis of the implication to RANSAC's hyperparameters and theoretical convergence expectations, and the experimental study of their effect 

\item a method for robust photo-realistic object reconstruction using NeRF and for robust single-shot multi-class reconstruction using LFNs
\item a qualitative/quantitative evaluation of our method on both synthetic and real-world data with different inconsistencies (occlusions, invalid calibrations, ...) indicating the state-of-the-art performance of our approach


\end{itemize}


\section{Related Work}
\label{sec:rel_work}
Among the vast literature on neural fields, the seminal work of Mildenhall et al. \cite{mildenhall2020nerf} opened many avenues in the computer vision community. 
It contributed to state-of-the-art solutions for novel view synthesis and 3D reconstruction that have been covered in respective surveys~\cite{Xie2021NeuralFI,tewari2022advances}. Noteworthy is the more recent contribution of instant neural graphics primitives (iNGP) \cite{mueller2022instant}, which uses a hash table of trainable feature vectors alongside a small network for representing the scene. iNGP achieved major run-time improvements, thereby enhancing the feasibility of practical applications for neural fields. 

Baseline models are highly sensitive to imperfections in the input data, which led to many works on robustness enhancements of neural fields; addressing a reduced amount of input views \cite{niemeyer2022regnerf, kim2022infonerf, wang2023sparsenerf, yu2021pixelnerf}, errors in camera parameters \cite{lin2021barf, jeong2021self, zhang2022relpose, bian2023nope}, variations in illumination conditions across observations \cite{brualla2021nerfw, sun2022neuconw}, multi-scale image data \cite{xiangli2022bungeenerf, barron2021mipnerf, barron2022mipnerf360}, and the targeted removal of floating artifacts \cite{warburg2023nerfbusters, wirth2023post, philip2023floaters}. 

Fewer works solve the reconstruction task in the presence of inconsistencies between observations. 
Bayes' Rays \cite{goli2023bayes} provides a framework to quantify uncertainty of a pretrained NeRF by approximating a spatial uncertainty field. It handles missing information due to self-occlusion or missing perspectives well, but cannot deal with inconsistencies caused by noise or distractors. Similarly, NeuRay \cite{liu2022neuray} only supports missing, but not inconsistent information.
Naive occlusion handling via semantic segmentation requires the occluding object types to be known in advance \cite{brualla2021nerfw, rematas2022urf, tancik2022block, rebain2022lolnerf, turki2022meganerf}. Solutions to learn semantic priors on transience exist \cite{lee2023semantic} but separating occlusions via semantic segmentation without manual guidance is ill posed.
Occ-NeRF \cite{zhu2023occlusion} considers any foreground element as occlusion and removes them via depth reasoning, but their removal leaves behind blurry artifacts. Alternatively, some methods do not remove dynamic distractors, but reconstruct them together with the rest of the scene using time-conditioned representations~\cite{liu2023robustdyn, wu2022d, pumarola2021d, chen2022hallucinated}. 
%
Closely related to our work, RobustNeRF~\cite{sabour2023robustnerf} considers input-image distractors as outliers of the model optimization task. The authors employ robust losses improved via patching to preserve high-frequency details. RobustNeRF does not rely on prior assumptions about the nature of distractors, nor does it require preprocessing of the input data or postprocessing of the trained model. Nevertheless, their method comes at the cost of loosing view-dependent details and a reduced reconstruction quality in undistracted scenes. Furthermore, their method is limited to unconditioned models that overfit to a single scene. A generalization to conditioned NeRFs, such as pixelNeRF \cite{yu2021pixelnerf}, is not obvious, as no further optimization takes place during inference.

Conditioned neural fields offer a distinct advantage in their ability to generalize to novel scenes by leveraging knowledge acquired from diverse scenes during learning. This results in a more robust model that requires as few as one input view for inference, showcasing the efficiency and adaptability of the approach. 
Contrary to PixelNeRF~\cite{yu2021pixelnerf}, which relies on a volumetric parametrization of the scene, demanding multiple network evaluations along the ray, Light Field Networks (LFNs)~\cite{sitzmann2021lfns}, which succeed Scene Representation Networks~\cite{sitzmann2019srns}, take a different approach. LFNs represent the scene as a 4D light field, enabling a more efficient single evaluation per ray for inference. The network takes as input a ray represented in Plücker coordinates and maps it to an observed radiance, all within an autoencoder framework used for conditioning.

None of the mentioned methods can deal with occlusions in single-shot reconstruction and no prior work exists on robust LFNs or robustness of other conditioned neural fields for single-shot reconstruction, which we address via the RANSAC paradigm \cite{fischler1981random}. Since its introduction in 1981, RANSAC has gained attention for fitting analytical models with a small number of parameters, such as homography estimation in panorama stitching \cite{brown2007panorama}. Among the few direct applications of classical RANSAC to larger models is robust morphable face reconstruction \cite{egger2016occlusion, egger2018occlusion}. Other common expansions and applications include differentiable RANSAC \cite{brachmann2017dsac} for camera parameter estimation in a deep learning pipeline, locally optimized RANSAC \cite{chum2003locally} to account for the requirement of a descriptive sample set, and adaptive real-time RANSAC \cite{raguram2008arrsac}.

\section{Method}
\input{figures/method}
In this section, we present our approach to increase the robustness of neural scene representations to inconsistencies in the input data.
First, we recap RANSAC, its theoretical convergence and hyperparameters, and the required adaptions for its application to high-dimensional data-driven models.
We then introduce a general scheme for the random sampling and validation of neural fields.
Based hereon, we formulate a robust algorithm using LFNs for 3D reconstruction from a single image with occlusions. Finally, using NeRF, we formulate an algorithm for robust photo-realistic reconstruction from multiple views in the presence of common sources of inconsistencies.

\subsection{RANSAC Convergence on Complex Models}
Classical RANSAC \cite{fischler1981random, choi1997ransacperformance, derpanis2010ransacoverview} follows an iterative process. Initially, a minimal set of samples is randomly selected to determine the model parameters, known as the hypothesis generation phase. Then, the hypothesis is evaluated by assessing the number of observations it explains, within a specified margin. These steps are repeated until the best hypothesis is chosen to constitute the consensus set, which comprises all of its inliers.

This paradigm cannot be directly applied to complex models such as neural fields, as a significant amount of samples is required to obtain decent initial model parameters and additional clean samples improve the quality further. This imposes a challenge regarding the expected amount of clean initial sample sets, $S_{clean}$:
\begin{equation}
  \abovedisplayskip=0.2cm
 \belowdisplayskip=0.2cm
\label{equ:expcleansamples}
    \mathbb{E}[\# S_{clean}] = N * \prod_{m=1}^{M} \frac{s_{img} - s_{img}^{occ} - m}{s_{img} - m},
\end{equation}
where $s_{img}$ denotes the total amount of samples (e.g., image pixels), $s_{img}^{occ}$ represents the occluded samples, and $N$/$M$ are the number of iterations/samples. The expected amount exponentially decreases with the initial number of samples. 

The samples and respective requirements for analytic and data-driven models vary a lot. The effect of individual samples is less traceable in data-driven models and the information entropy varies more significantly across samples. When using a model that projects onto a latent space, some very atypical outliers do not show an effect at all if the latent space is not expressive enough to explain them in an overall loss-reducing way, yet, outliers close to the object or its color, or larger chunks of outliers, will usually be distracting. At the same time, samples of small-scale high-frequency 
details are important for the reconstruction and contain a lot of information, whereas multiple samples of larger-scale lower-frequency details contribute much less. The amount of initial samples for the hypothesis generation becomes a tunable hyperparameter trading initial reconstruction quality for likelihood of finding desired sample sets. This invalidates the classical convergence idea \cite{choi1997ransacperformance, derpanis2010ransacoverview} where the RANSAC iterations $N$ with
\begin{equation}
  \abovedisplayskip=0.2cm
 \belowdisplayskip=0.2cm
\label{equ_iter_from_exp}
    N \ge \frac{\log(1-p)}{\log(1-t^M)}
\end{equation}
are chosen such that at least one clean sample set is found with a probability $p$, given the expected ratio of clean samples $t$ and the amount of initial samples $M$. There is not only the need to find a clean sample set, but one that captures all important details. At the same time, a completely clean sample set is not required at all, as long as the contained outliers are not represented by the local minimum of the latent space or the model itself, depending on the concrete scene representation.

\subsection{Random Sampling Neural Fields}
We propose a general strategy for RANSAC-like, iterative, robust reconstruction with neural fields, before formulating respective algorithms for LFNs and NeRF.

\noindent\textbf{Sampling Hypotheses}\quad Based on the application a feasible sampling domain is chosen \eg pixels/rays, or observations. 
%
Depending on the requirements of the neural field an appropriate sample size is determined. The initial sample size is chosen as small as possible to allow for a reasonable convergence expectation to cleaner sets, while being large enough for a coarse fit of the model expressive enough to enable the discrimination of outliers. 

\noindent\textbf{Determine Model Parameters}\quad The determination of the model parameters in a classical RANSAC application corresponds to the inference of the sample sets with a neural field: In case of unconditioned neural fields (such as NeRF) this corresponds to (over-)fitting the model to the scene, in case of conditioned fields this corresponds to obtaining a latent representation. Full convergence is not required, the reconstruction is only used to evaluate the outlier contamination.

\noindent\textbf{Validation of Hypotheses}\quad For the validation, the rays/perspectives not used for the inference are rendered using the obtained model and compared to the input. The quality of each hypothesis is evaluated based on the amount of other samples explained by it up to some margin. Depending on the sampling/inference strategy, different similarity metrics can be used to distinguish outliers from other sources of noise such as the coarse inference. The initial sample set of the best hypothesis together with all its inliers is used for a final complete model fit.

As will be demonstrated in our evaluation in Section \ref{sec:experiments}, the major benefit of our RANSAC-like neural field approach over formulations based on robust loss functions \cite{barron2019general} is the possibility to reliably filter outliers and inconsistencies in the input data in comparison to the down-weighting of their influence.

\subsection{Robust Light Field Networks}
In the following, we propose a novel fast and robust single-shot multi-class reconstruction algorithm  based on LFNs \cite{sitzmann2021lfns}. LFNs are globally conditioned, meaning that the supported subset of 3D consistent scenes is represented by a single global latent vector.
Therefore, when the latent space is not expressive enough to represent object and distractor correctly, large inconsistencies cause global damage instead of local artifacts.
%
LFNs intrinsically support parallel inference, allowing to jointly process all hypotheses rather than following an iterative scheme. Our algorithm consists of the following steps (Fig. \ref{fig:ranrac_overview}):

\begin{enumerate}[leftmargin=\parindent]
 \item {\textbf{Hypothesis Consensus Set:} Given the input image $I$ as a set of pixel color values $\myVector{c}_i$, and the intrinsic and extrinsic camera parameters, the set of rays $R$ -- one ray $\myVector{r}_i$ for every pixel -- represented by Pl\"ucker coordinates, is generated. In the first step, $N$ initial consensus sets $S_n$ are drawn using a uniform distribution, where each consensus set consists of $M$ random samples:
 \begin{equation}
     \setlength{\belowdisplayskip}{0pt} \setlength{\belowdisplayshortskip}{0pt}
    \setlength{\abovedisplayskip}{0pt} \setlength{\abovedisplayshortskip}{0pt}
    (\myVector{c}_n^m, \myVector{r}_n^m) \in_R \{(\myVector{c}_i, \myVector{r}_i) \mid \myVector{c}_i \in I, \myVector{r}_i \in R\}
 \end{equation}
 where $n \in\{1,\dots,N\}$, $m \in \{1, \dots, M\}$, and $\in_R$ denotes a sample randomly drawn from the set without replacement according to a uniform distribution.
 
 }
 \item {\textbf{Hypothesis Inference:} The autodecoder of an LFN $\Phi$ with hypernetwork $\Psi$ and pretrained hypernetwork weights $\psi$, is used to infer the latent codes $\myVector{z}_n$ for each of the initial sample sets in parallel. 
 \begin{equation}
     \setlength{\belowdisplayskip}{0pt} \setlength{\belowdisplayshortskip}{0pt}
    \setlength{\abovedisplayskip}{0pt} \setlength{\abovedisplayshortskip}{0pt}
    \{\myVector{z}_n\} = \argmin_{\{\myVector{z}_n\}} \sum_n\sum_m 
    \lVert
    \Phi(\myVector{r}_{n}^m \mid \Psi_\psi(\myVector{z}_n)) - \myVector{c}_n^m
    \rVert_2^2
    + \lambda_{lat} \lVert \myVector{z}_n \rVert_2^2
 \end{equation}
 $\lambda_{lat}$ determines the strength of the Gaussian prior on the latent space \cite{sitzmann2021lfns}. An exponential learning rate schedule speeds up the inference. The inferred latent codes form the hypotheses. 
 }
 
 \item {\textbf{Hypothesis Prediction:} Each of the hypotheses is used to render an entire image $I_{n}^{pred}$ from the perspective of the input image, each consisting of the pixel color values
$
     \myVector{c}_{n,i}^{pred} = \Phi(\myVector{r}_{i} \mid \Psi_\psi(\myVector{z}_n))
$
, using again the set of rays $R = \{\myVector{r}_i\}$ obtained from the camera parameters. The rendered pixels resemble the predictions for the remaining observations under each hypothesis. 
 }
 
 \item {\textbf{Hypothesis Validation:} The obtained predictions are compared to the input image to validate the hypothesis. For each pixel in each predicted image, we calculate the Euclidean distance in color space:
$
    e_{n,i} = \lVert \myVector{c}_{n,i}^{pred} - \myVector{c}_i \rVert_2.
$
    For each image, using these distances, we collect, up to some margin $\epsilon$, the observations explained by the model (inliers) : 
$
    S_n^{inlier} = \{(\myVector{c}_i, \myVector{r}_i) \mid e_{n,i} < \epsilon\}
$
}

 \item {\textbf{Model Selection:} We select the best hypothesis sample set $S_{best}$ based on the number of inliers $\# S_n^{inlier}$.
The model is inferred once more, similar to the second step, to obtain the final latent code $z_{cons}$. The inference is based on the final consensus set $S_{cons} = S_{best} \cup S_{best}^{inlier}$, the initial sample set of the strongest hypothesis $S_{best}$ together with all its inliers $S_{best}^{inlier}$.
}
The final output consists of both the latent code $z_{cons}$ and the final consensus set $S_{cons}$ of the selected model, and can be used to render arbitrary new perspectives.
\end{enumerate}

\subsection{Robust Neural Radiance Fields}
%
In the following, we propose a novel robust multi-view reconstruction approach based on NeRFs.
NeRFs are fit to a single scene based on a set of observations. As they support view-dependent radiance, one might expect inconsistencies in the input observations to only have an effect on specific perspectives. However, the density is only spatially parametrized, hence, inconsistencies lead to significant ghosting and smearing artifacts in more than just the inconsistent perspective. This can be leveraged in the hypothesis validation. The following steps are performed for N iterations:

\begin{enumerate}[leftmargin=\parindent]
    \item {\textbf{Hypothesis Consensus Set:} A NeRF fit requires a large set of rays, making a sampling in ray space infeasible, as even obtaining significantly clean\textit{er} sets becomes unlikely. In order to obtain a sampling domain with reasonably sized initial consensus sets $S_n$, we sample $M$ observations from the given sets of images $\mathcal{I}$ and corresponding camera poses $\mathcal{C}$ in every iteration:
    \begin{equation}
    \setlength{\belowdisplayskip}{0pt} \setlength{\belowdisplayshortskip}{0pt}
    \setlength{\abovedisplayskip}{0pt} \setlength{\abovedisplayshortskip}{0pt}
            (I_n^m, E_n^m) \in_R \{(I_i, E_i) \mid I_i \in \mathcal{I}, E_i \in \mathcal{C}\}
    \end{equation}
     where $n \in\{1,\dots,N\}$, $m \in \{1, \dots, M\}$, and $\in_R$ denotes a sample randomly drawn from the set without replacement according to a uniform distribution.
    }
    \item {\textbf{Hypothesis Inference:} The sampled observations are used to fit a hypothesis neural radiance field $F_n$}.
    \item {\textbf{Hypothesis Prediction:} The obtained NeRF $F_n $ is used to render predictions  $I^{pred}_{n, i}$ with pixel colors $\myVector{p}_{n, i, x}^{pred}$ under the hypothesis
    for all unseen input perspectives $E_i$.
    }
    \item {\textbf{Hypothesis Validation:}
    The chosen sample space requires a careful evaluation of the hypotheses. We propose a two-step evaluation, where, first, the pixels inliers $P_{n, i}$ up to some margin $\epsilon_{pix}$ are determined for every observation based on the Euclidean distance in color space to the pixels $\myVector{p}_{i, x}$ of the input images $I_i$:
    $
        P_{n,i} = \{ \myVector{p}_{n, i, x}^{pred} \mid 
        \lVert \myVector{p}_{n, i, x}^{pred} - \myVector{p}_{i,x} \rVert_2
        < \epsilon_{pix}
        \}
    $
    and, second, the observations themselves are labeled as inliers or outliers based on the amount of pixel inliers, again up to some margin $\epsilon_{img}$, to obtain the consensus set $S^{inlier}_n$ under the hypothesis:
    $
    S^{inlier}_n = \{(I_i, E_i) \mid \#P_{n,i} > \epsilon_{img}\}
    $.
    The binary metric for pixels ensures that smaller mispredictions (due to, e.g., view-dependent lighting effects) do not introduce noise into the evaluation.
    }
    \item {\textbf{Model Selection:}
     The strongest hypothesis, with its initial sample set $S_{best}$, is selected based on the number of inliers $\# S_n^{inlier}$ to obtain the final consensus set $S_{cons} = S_{best} \cup S_{best}^{inlier}$. The final model is obtained with one more NeRF fit of the consensus set.}
\end{enumerate}

\subsection{Hyperparameters}
For LFNs, the amount of initial samples and random hypotheses to evaluate (iterations), are determined experimentally. Without fine-tuning per class, the experimentally determined parameters are 90 initial samples and 2000 iterations, which supersedes the theoretical value for a  convergence because the latent space introduces an intrinsic robustness.  
The inlier margin balances the amount of slight high-frequency variations that are being captured and the capability of separating outliers that are similar to the object. A margin of 0.25 in terms of the Euclidean distance of the predicted colors to the input samples in an RGB color space normalized to the range $(-1, 1)$ has been found to be optimal. 
Please refer to the supplementary material for further experimental results.

For NeRFs, the parameters behave more natural and the amount of perspectives required for a meaningful, not completely artifact-free fit of the model lies around 25 observations \cite[Table~2]{mildenhall2020nerf}. With fewer samples, more artifacts are introduced that get harder to separate from the ones caused by inconsistencies, and the samples get more dependent on being evenly spaced. For real-world captures with 10\% inconsistent perspectives, as few as 50 iterations are sufficient. 
With the color space normalized to $(0, 1)$, a pixel margin between around 0.15 in terms of Euclidean distance worked well for the determination of actual artifacts. We consider an observation an inlier based on a margin of 90\% - 98\% pixel inliers, which proved to be a good choice to separate minor artifacts (due to the sparse sampling) from artifacts caused by actual inconsistencies. For different datasets or inconsistencies, these values can be adapted.

\section{Implementation \& Preprocessing}
For LFNs, we build on top of the original implementation \cite{sitzmann2021lfns}, with a slight adaptation to enable a parallel sub-sampled inference. We furthermore use the provided pretrained multi-class model. The camera parameters are known. For efficiency reasons, the steps of the algorithm are not performed iteratively, but multiple hypotheses are validated in parallel. To further speed up the inference, an exponential learning rate schedule is used for the auto-decoding, leading to a total runtime of about a minute on a single GPU.

For the robust reconstruction of objects from lazily captured real-world data, one has to estimate the camera parameters and extract foreground masks before applying the algorithm. For the estimation of the camera parameters, we used the COLMAP structure-from-motion package \cite{schoenberger2016sfm}. We extracted foreground masks using Segment Anything \cite{kirillov2023segany}. However, only foreground masks, containing the objects and the occlusions, are extracted. Segment Anything is not capable of removing arbitrary occlusions in an automatized way. 
After these preprocessing steps, the robust reconstruction algorithm can be applied as described. Erroneous estimates of the camera parameters or foreground masks are excluded by our algorithm, thus making the entire reconstruction pipeline robust.
 
Our algorithm is not limited to a specific NeRF implementation. The chosen sampling domain eases integration into arbitrary existing NeRF implementations, which commonly expect images instead of unstructured ray sets.
However, using a fast NeRF variant is advantageous when applying an iterative scheme. We used the instant NGP implementation \cite{mueller2022instant} of the instant NSR repository \cite{instant-nsr-pl}, which includes some accelerations \cite{tiny-cuda-nn, li2023nerfacc}. Other, (specifically fast) variants are likely good choices as well. Antialiased and unbounded, but slow variants, such as MipNeRF360 \cite{barron2022mipnerf360}, are not feasible. For further implementation details please refer to the supplementary.

\section{Experiments}\label{sec:experiments}
\subsection{Inconsistencies, Baselines \& Datasets}
\noindent\textbf{Multi-View Reconstruction (NeRF)}\quad We provide a comprehensive quantitative and qualitative analysis of RANRAC's multi-view reconstructions compared to NeRF without any method of robustness (baseline) and RobustNeRF \cite{sabour2023robustnerf} (state of the art). We conduct experiments for various common sources of inconsistencies such as occluded perspectives, noisy camera parameter estimates, and blurred perspectives.
RobustNeRF \cite{sabour2023robustnerf} targets unbounded scenes with multiple objects and small amounts of distractors in every perspective. In comparison, our RANSAC-based approach deals well with single-object reconstruction, even with heavy occlusions, as long as enough clean perspectives are available. As their dataset reflects the algorithm's properties, we cannot provide a fair comparison. Instead, we demonstrate the applicability using a custom dataset of a single object with a controlled amount of deliberately occluded perspectives. For the other inconsistencies we use of-the-shelf datasets and add noise to the camera parameters and blur to the images.
Furthermore, we implement the robust losses \cite{sabour2023robustnerf} on the same NeRF variant as RANRAC to provide a fair comparison of the method's robustness, independent of the NeRF variant. For further details on the implementation, please refer to the supplementary material.



\noindent\textbf{Single-Shot Reconstruction (LFN)}\quad
We benchmark against the original LFN implementation of Sitzmann et al. \cite{sitzmann2021lfns} as baseline, as there are no other robust methods for LFNs or  conditioned neural fields, nor are there methods for robust single-shot multi-class reconstruction in general. Furthermore, we use the same pretrained LFN for the baseline and for the application of our method. The LFN is pretrained on the thirteen largest ShapeNet classes~\cite{chang2015shapenet}.

We provide a detailed qualitative and quantitative performance comparison under different amounts of occlusion for three representative classes (plane, car, and chair), while just stating reconstruction performance in a fixed environment without additional tuning of the hyperparameters for the others. The plane class is mostly challenging due to the low-frequency shape, while the car class contains a lot of high-frequency color details. The chair class represents shapes that are generally problematic for vanilla LFNs, even without occlusions. We provide a complementing analysis of the hyperparameters in the supplementary material. If not stated otherwise, we evaluate using 50 randomly selected images of the corresponding class. All comparisons use the same images. 

The occlusions are created synthetically. They consist of randomly generated patches while controlling two metrics of occlusion: Image occlusion and object occlusion. The former is the naive ratio of occluded over total pixels. The latter are the occluded pixels on the object compared to the total pixels covered by the object. We use both metrics to take the vastly different information entropy of samples across the image into account
For further details on the generation of the synthetic occlusions, please refer to the supplemental.

\subsection{Evaluation}
\input{figures/lfn_table}
\input{figures/lfn_plot_imgocc}
\input{figures/lfn_quali_combined}
\input{figures/nerf_qual_new}
\input{figures/nerf_table_combined}
For LFNs, our approach leads to a significant improvement in occluded scenarios of up to 8dB in PSNR and a similarly strong improvement for the SSIM. The improvement is most significant in heavily distracted scenarios (Fig. \ref{fig:recransacoccimg}). In clean scenarios a slight performance penalty can be observed, but even with small amounts of object occlusion (information loss), our algorithm outperforms the baseline, leading to numerically better results up to $50\%$ information loss (Fig. \ref{fig:recransacoccimg}).
The effect is not only measurable, but also well visible (Fig.~\ref{fig:lfn_quali_combined}). Increasing amounts of occlusion slowly introduce local artifacts into our reconstruction while preserving a reasonable shape estimate even for larger amounts of occlusion. In contrast, the reconstruction of LFNs breaks rather early in a global fashion. Still, our consensus set (Fig.~\ref{fig:lfn_quali_combined}), reveals that some high-frequency details were wrongfully excluded, explaining the slight performance decrease on clean images.
In general, the benefit of RANRAC is best observable for classes that can be well described by LFNs, as evidenced by the lower improvements the chair class exhibits, compared to the plane and car class. The same effect is visible in the quantitative evaluation on the other classes without additional tuning (Tab. \ref{tab:quant_other_classes}).
The only outlier is the display class, on which LFNs struggle the most on \cite{sitzmann2021lfns}, even in unoccluded scenarios. It is not consistently represented in the latent space and robust reconstruction amplifies this effect by reducing the samples to a consistent set.


The application to NeRF shows the significantly improved reconstructions of RANRAC (Tab.~\ref{tab:nerf_eval}) for different types of inconsistencies such as occlusions and blur, as well as noisy camera parameters on captured and synthetic datasets. RANRAC consistently outperforms RobustNeRF \cite{sabour2023robustnerf} for object reconstruction from inconsistent inputs. RobustNeRF specifically struggles in capturing view-dependent appearance and at concavities while RANRAC seamlessly reconstructs them. The effects are not only measurable but also well-visible (Fig.~\ref{fig:nerf_floaters_removed}).

\section{Limitations \& Future Work}
For the application to NeRF, the iterations imply the use of a fast inferable variant, ruling out MipNeRF360 \cite{barron2022mipnerf360} and other high-quality variants for unbounded scenes. Further, foreground separation is a must, limiting this application of RANRAC to single-object reconstruction. The recent 3D Gaussian Splatting~\cite{kerbl2023gaussians} raises optimism that this limitations may soon be lifted.

By following a RANSAC-like approach, we inherit the requirement of sufficient clean perspectives, which could be lifted via NeRF variants that require fewer perspectives \cite{niemeyer2022regnerf,truong2023sparf} or by using different sampling domains. 
In return, our method is not limited to specific kinds of inconsistencies, and is robust to arbitrarily heavy distractions or inconsistencies in the impure perspectives.

Wheras our robust LFN approach led to significant improvements for single-shot reconstruction, our concept might also be applied to other conditioned neural fields, based on a smart choice of the sampling domain, targeting photo-realism. One could also use importance sampling based on a prior instead of a uniform sampling, leveraging the unevenly distributed information entropy. Neural sampling priors \cite{brachmann2019ngransac}, semantic segmentation and more advanced schemes (e.g. locally-optimized RANSAC \cite{chum2003locally} or DSAC \cite{brachmann2017dsac}) might also prove useful.

\section{Conclusion}
We introduced a novel approach to increase the robustness of neural fields, inspired by the RANSAC paradigm.
Following this concept, we introduced a novel robust approach for single-shot reconstruction from occluded views based on LFNs which achieves a significant improvement in reconstruction quality for distracted and occluded scenarios, even for extreme cases.
%
Furthermore, we introduced a respective RANRAC-based NeRF variant that allows robust photo-realistic reconstruction from multiple views with typical inconsistencies such as occlusions, noisy camera parameters, or blurred images - resulting in significant improvements compared to state-of-the-art methods - without relying on assumptions about the distractions.


\section*{Acknowledgements}
Benno Buschmann was at FAU funded through a gift by Mitsubishi Electric Research Laboratories (MERL). Andreea Dogaru was funded by the German Federal Ministry of Education and Research (BMBF), FKZ: 01IS22082 (IRRW). The authors are responsible for the content of this publication. 

\bibliographystyle{splncs04}
\bibliography{egbib}

\end{document}


\title{RANRAC: Robust Neural Scene Representations via Random Ray Consensus
\\
Appendix}

\titlerunning{RANRAC Appendix}


\author{
Benno Buschmann \inst{1,2} \and
Andreea Dogaru \inst{2} \and
Elmar Eisemann \inst{1} \and
Michael Weinmann \inst{1} \and
Bernhard Egger \inst{2}
}

\authorrunning{B.~Buschmann et al.}


\institute{Delft University of Technology, Netherlands \and
Friedrich-Alexander-University Erlangen-Nürnberg, Germany}


\maketitle

\section{Overview}
In the scope of this supplemental, we provide more details regarding our implementation of RANRAC for LFNs~\cite{sitzmann2021lfns} and NeRFs~\cite{mildenhall2020nerf}, as well as the implementation of RobustNeRF~\cite{sabour2023robustnerf} used for benchmarking. Furthermore, we present more information on the RANRAC hyperparameters for the application to LFNs and NeRFs. Finally, we provide a more detailed qualitative evaluation that demonstrates the potential of our approach with respect to different types of inconsistencies for different scenes.

\noindent Project Page: \url{https://bennobuschmann.com/ranrac/}

\section{RANRAC for LFNs}
\subsection{Implementation Details}
For the inference with LFNs we used 500 iterations with a learning rate of 0.001 and an exponential schedule. Regarding all other aspects, we follow the implementation published by the authors of LFNs~\cite{sitzmann2021lfns}.
The integration of the parallel inference of the different sample sets allows for a faster inference of the randomly sampled hypotheses, as the individual hypotheses, their inference, and their evaluation do not depend on each other. On an A100 GPU with 40 GB of RAM, all 2048 hypotheses can be inferred in parallel resulting in a total runtime of about a minute. We also performed tests on an Nvidia RTX 3060 GPU with 12 GB, where the inference is split into batches of 512 hypotheses.

\subsection{Experimental Hyperparameters}
We provide experimental results regarding the effect of different hyperparameter choices of RANRAC by varying one parameter and keeping all others fixed to the designated optimum to observe the isolated influence. If not specified otherwise, the reported results are based on 50 randomly selected images per class. We used reasonable amounts of occlusion between 20\% and 30\%.

\textbf{Effect of the number of initial samples} --
As discussed, the amount of initial samples taken is a trade-off between the probability of finding enough clean(er) sample sets, and the reconstruction quality from those samples. In Figure \ref{fig:reclfnsamples}, we provide insights on the relationship between the reconstruction quality and the number of clean samples.
As expected, quality increases with more clean samples. Furthermore, the number of samples required for meaningful reconstructions varies between datasets, which is explained by the average amount of high-frequency details of a certain class, as well as its general complexity, and the strength of LFNs on a specific class. Furthermore, we observe that the quality greatly depends on where the samples land, as illustrated by the 99th percentile. This supports our finding that not only a sufficiently clean sample set but also one leading to meaningful reconstructions has to be found.

Fig. \ref{fig:samplesparamransac} shows the reconstruction performance of RANRAC for different classes and numbers of samples while keeping the other parameters (i. e. margin, iterations, amount of inconsistency) fixed.
It is well visible that there is not a single optimum across classes. For the plane class, a small sample size is enough. The quality is decent for $\sim$40-120 samples, it drops quite drastically above that limit. In contrast, the chair class needs more samples, i.e. $\sim$60-280 samples, and the quality even keeps slightly improving. The car class lies in between, i.e., at $\sim$60 samples a decent performance is reached and the performance still slightly, but not steadily, improves until $\sim$160 samples, and then slowly decreases again.
The difference is explained by attributes of the classes. The chair class contains a lot of zoomed-in observations, closer to the object, opposed to the plane and the car class. This results in more high-frequency details which need to be captured. LFNs have difficulties reconstructing the chair class, even in clean environments. Furthermore, the object covers a larger part of the image compared to the other classes. Not all occlusion attributes can be fixed at once. Although, object occlusion and image occlusion are controlled, the occlusion-to-object ratio is much lower on average for the chair class (28\%) compared to car (35\%) and plane (62\%). Thereby, an additional sample is more likely to be clean and beneficial for the former two classes.

As the reconstruction quality is already degenerate for plane, and stagnating for car, we did not perform tests with more than 280 samples.
When tuning per class is not feasible, we propose choosing 90 samples. We used this for the evaluation of all classes in the main paper. The choice is within the identified optimal range of the plane and car class, and the performance penalty for the chair class is not too significant (less than 1dB).  A choice on the lower end of feasible options for the examined amount of occlusion has additional benefits: Taking fewer samples reduces run-time and is desirable for environments with a higher amount of occlusions. We, however, do not claim that this choice is optimal for all classes and environments.

\textbf{Effect of the margin} --
The selection of a reasonable choice for the inlier margin is bound by two factors. 
For tight margins, some clean samples are wrongfully excluded due to small high-frequency variations that are not represented by the initial down-sampled reconstruction. If it is too loose, some occluded samples are no longer separated. In Fig.~\ref{fig:qualmargineffect}, we show the consensus sets generated by RANRAC for different margins. Note that it is expected to have some occluded samples being explained by the model.

%
In Fig.~\ref{fig:recransacmargin}, we compare the reconstruction quality of RANRAC for different margins. A maximum reconstruction performance is observed at a margin of 0.25 in terms of Euclidean distance in a color space normalized to $(-1, 1)$. It is further observed that the plane class is more resilient to a lower margin than the others and the SSIM even starts reducing at a margin of 0.15, most likely due to the lower amount of high-frequency variations within this class. As the looser margin is not required to include any of those details, the inclusion of slight amounts of occlusion caused by it is visible earlier.

\textbf{Effect of the number of iterations} --
As expected, the quality improvement via additional random hypotheses (iterations) behaves approximately logarithmic. Powers of two are generally desirable for the parallel inference. We settled on 2048 iterations, as there is a significant improvement compared to 1024 iterations, and no improvement with 4096 iterations. Any measured decrease with more iterations can only be caused by noise.

\input{suple_aux/hp_additional_results}
\input{suple_aux/center_point_plot}

\subsection{Generation of Occlusions}
For the ShapeNet data \cite{chang2015shapenet}, the occlusions need to be simulated in a randomized but controllable way, and they have to represent the attributes of real-world occlusions. Variation in position and color is required, while the shape is not important, as our algorithm randomly draws the samples according to a uniform distribution, and does not rely on semantics. However, it is the most challenging environment to have a single connected patch of occlusion, as the information loss about the occluded object part is less curable. As this is also the most common type of occlusion under real-world conditions, we particularly focus on demonstrating the robustness of our approach to this scenario.
Furthermore, using a monotone color as occlusion would oversimplify the problem and allow for unrealistically high choices for the margin.
The reason is that a single monotone color is fairly easy to separate from the object, as long as it is not the same color as the object. Therefore, we occlude with noise patches, where each color channel is sampled separately from a normal distribution with decently large standard deviation. This results in randomly varying color across the patch and presents a bigger challenge, as it introduces distractions more similar to the object and more likely to be embedded in a local optimum of the latent space.
The final challenge is the placement of the occlusions. In addition to being randomized, the placement should model the common type of partial occlusion.
This reduces noise in the results due to completely occluded objects which are just unrecoverable. Furthermore, simulating the occlusion by smaller patches which just cover the object itself would oversimplify the problem, as the shape would still be well reconstructable. The correct way of simulating the occlusion is therefore to place an occlusion patch at about the edge of the object, which has the right size to cover the desired relative part of the object. This still has to be done in a randomized way.
A good practical realisation is drawing the coordinates of the center point of the occlusion patch independently from a distribution that consists of two symmetric Gaussian spikes at about the expected object borders (Figure \ref{fig:illustration_center_distribution}). Their mean is used to control the desired amount of object occlusion.

The size of the patch is drawn from independent normal distributions for the size in both dimensions, the mean is used to control the desired amount of image occlusion. Finally, the accomplished relative occlusion of the object is measured, and if it does not fall into the specified bandwidth of object occlusion, e.g. 20\%-30\%, a new patch is generated. This could be considered as rejection sampling of the desired distribution.


\section{RANRAC for NeRFs}
\subsection{Implementation Details}
When applying RANRAC in the context of NeRF-based reconstruction, we use the Instant-NSR implementation~\cite{instant-nsr-pl} with Adam optimizer, a learning rate of 0.01, a distortion loss lambda of 0.001, and a multi-step learning rate schedule with gamma 0.33. The parameters vary for the RANRAC hypothesis evaluation, where a shallow fit is adequate, and the final inference. Within the hypothesis evaluation, 4000 training iterations are sufficient; for the final inference, 20,000 training iterations are used. The hash grid is reduced to a maximum size of $2^{18}$, to ensure a decent performance with a wider range of GPUs (due to cache alignment \cite{mueller2022instant}). We use mixed precision training. Regarding the remaining parameters of the geometry and texture representation, we used the default parameters used by the authors of Instant-NSR. On acceptance, we will publish all configuration files together with the source code to provide all of the details and facilitate reproducing our results. The run-time is about a minute per iteration and less than five minutes for the final inference on an Nvidia RTX 4090 GPU. 

Regarding the RANRAC hyperparameters, a pixel margin of $\epsilon_{pix} = 0.15$ in terms of Euclidean distance in a color space normalized to $(0, 1)$ worked well for the determination of actual artifacts. We consider an entire observation as an inlier based on a margin of 90\%-98\% of its pixels being inliers, which proved to be a good choice to separate minor artifacts (due to the sparse sampling) from artifacts caused by actual inconsistencies. For the dataset with less heavy occlusion, 50 iterations were sufficient. For the dataset with heavier occlusion, we reported results after 500 iterations in the main paper. But even after less than 100 iterations, a significantly improved quality can already be noted. For other inconsistencies we evaluated with 100 iterations.

\subsection{Further Implementation Details regarding Baseline (RobustNeRF) for Benchmarking}
To allow a fair, isolated comparison of the robustness method, we implemented the losses introduced by RobustNeRF~\cite{sabour2023robustnerf} on top of the Instant-NSR method~\cite{instant-nsr-pl} used for our implementation. Even though the authors of RobustNeRF claim the generality of their method, some practical challenges and incompatibilities hamper the transfer. We resolved any appearing conflict as natural as possible and actively refined their method where it was necessary to achieve a decent performance in the provided scenario and allow a decent integration into fast NeRF variants. The details are presented in the following paragraphs.

\textbf{Patch-based Sampling} -- A key component of the robust losses is the patch-based sampling to evaluate the inliers in a neighbourhood. This leads to a conflict with the occupancy grid used by Instant-NSR~\cite{instant-nsr-pl} or iNGP \cite{mueller2022instant}. When only updating a small amount of patches, compared to thousands of randomly distributed locations, the occupancy estimate naturally becomes more biased and less accurate. Depending on the sampled patches, it can happen that none of them are considered as occupied according to the grid, and, therefore, no gradient is available and no update takes place.
This can lead to divergence without recovery. We selected only converging runs of RobustNeRF when benchmarking.

Furthermore, the amount of patches used as reported by the robustNeRF authors is not feasible for tiny-cuda-nn, we had to reduce it from 64 to 16 to allow the model to fit into the 24GB VRAM of an Nvidia RTX 4090 GPU, implying a slightly more stochastic gradient descent.

\textbf{Remark on Mask Computation} -- The authors mention in the paper that the patch mask is computed based on the smoothed neighbourhood inlier mask of the second step \cite{sabour2023robustnerf}. However, in the source code they published \cite{RobustNeRFCode}, the patch mask is computed based on the per-pixel inlier mask of the first step. The difference is naturally rather negligible. We still considered it worth noting, and report that we applied it as provided by their implementation.

\textbf{Single object reconstruction} --
When directly applying the method as described in the paper to the single-object reconstruction scenario, a reconstruction of the white background is obtained, and the object is wrongfully considered as the distraction and removed. For that reason, we had to use a higher margin of 0.8 instead of median, to allow their method to work at all for this task.

Finally, we used mixed precision for ours and full precision for theirs, as with their method, sometimes no sample is considered as being occupied and an inlier, and therefore no gradient update is possible. This does not play well with mixed precision in pytorch lightning. If at all, this difference benefits their method.

\subsection{Details regarding the Occluded Watering Pot Dataset}
\begin{figure}[!htb]
\begin{tabular}{cc}
    \setlength\tabcolsep{0pt}
    \renewcommand{\arraystretch}{0}
    \centering
    \begin{tabular}{cc}
    \includegraphics[width=0.23\textwidth]{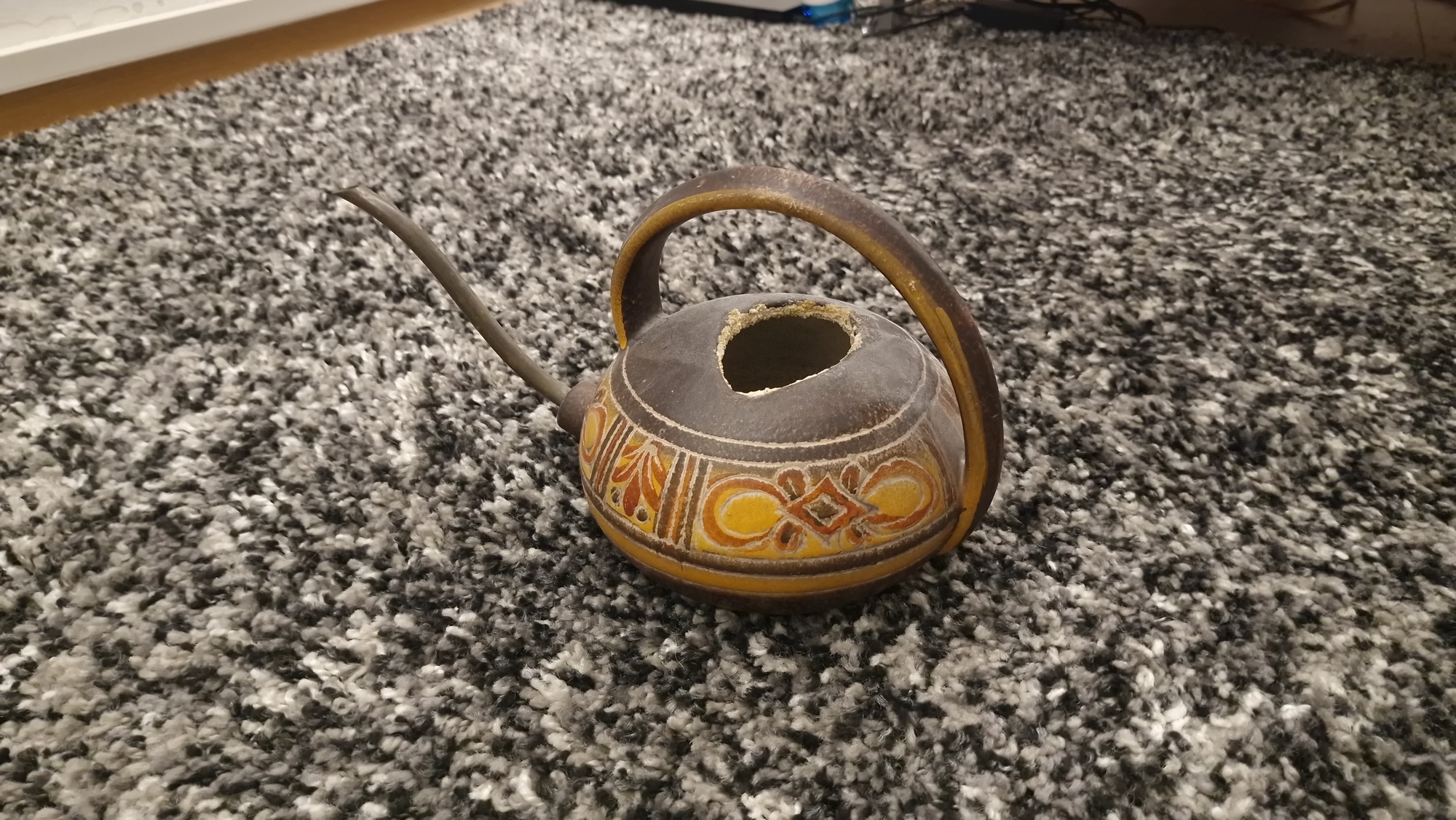}
    &
    \includegraphics[width=0.23\textwidth]{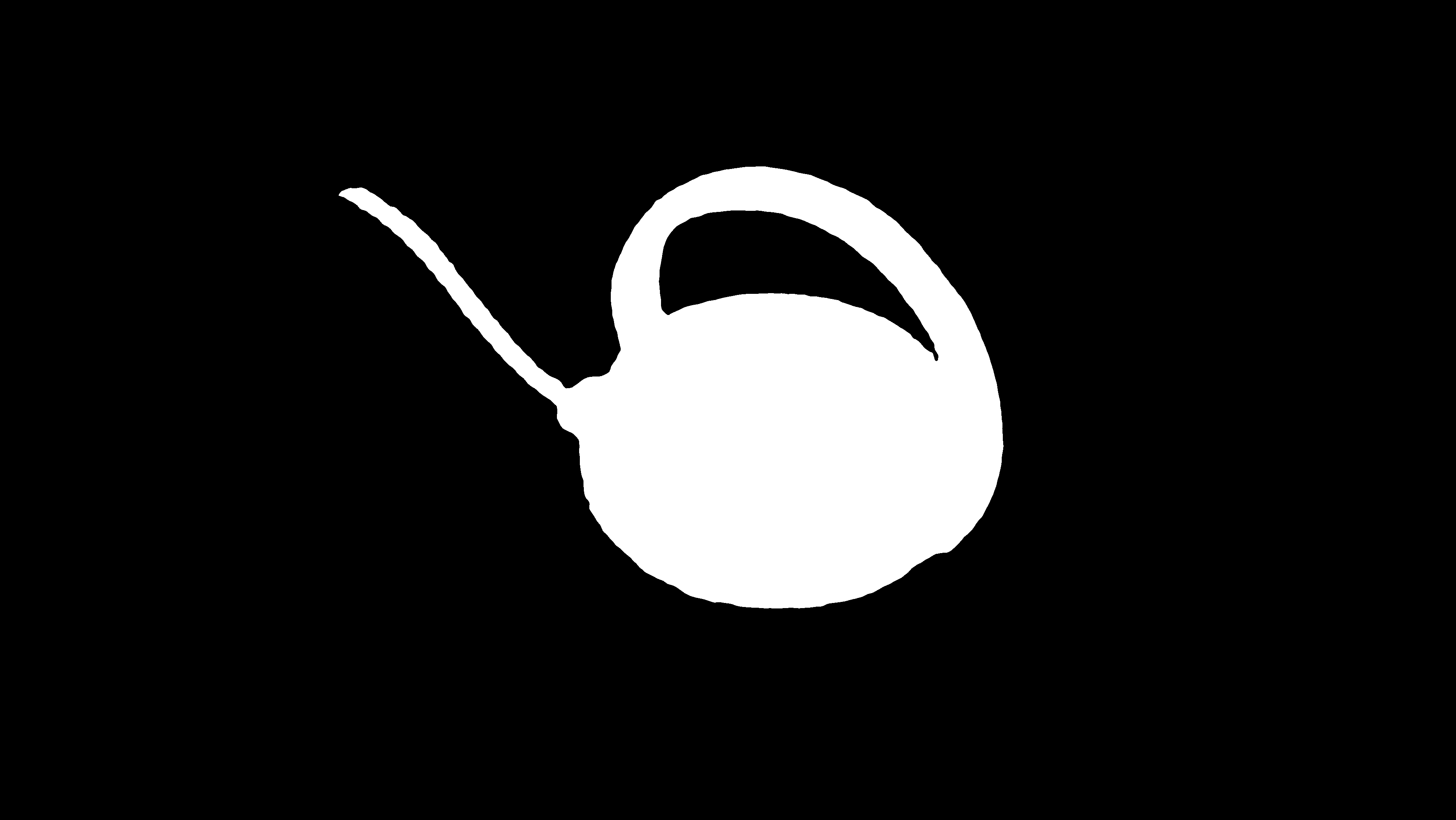}
    \\
    \includegraphics[width=0.23\textwidth]{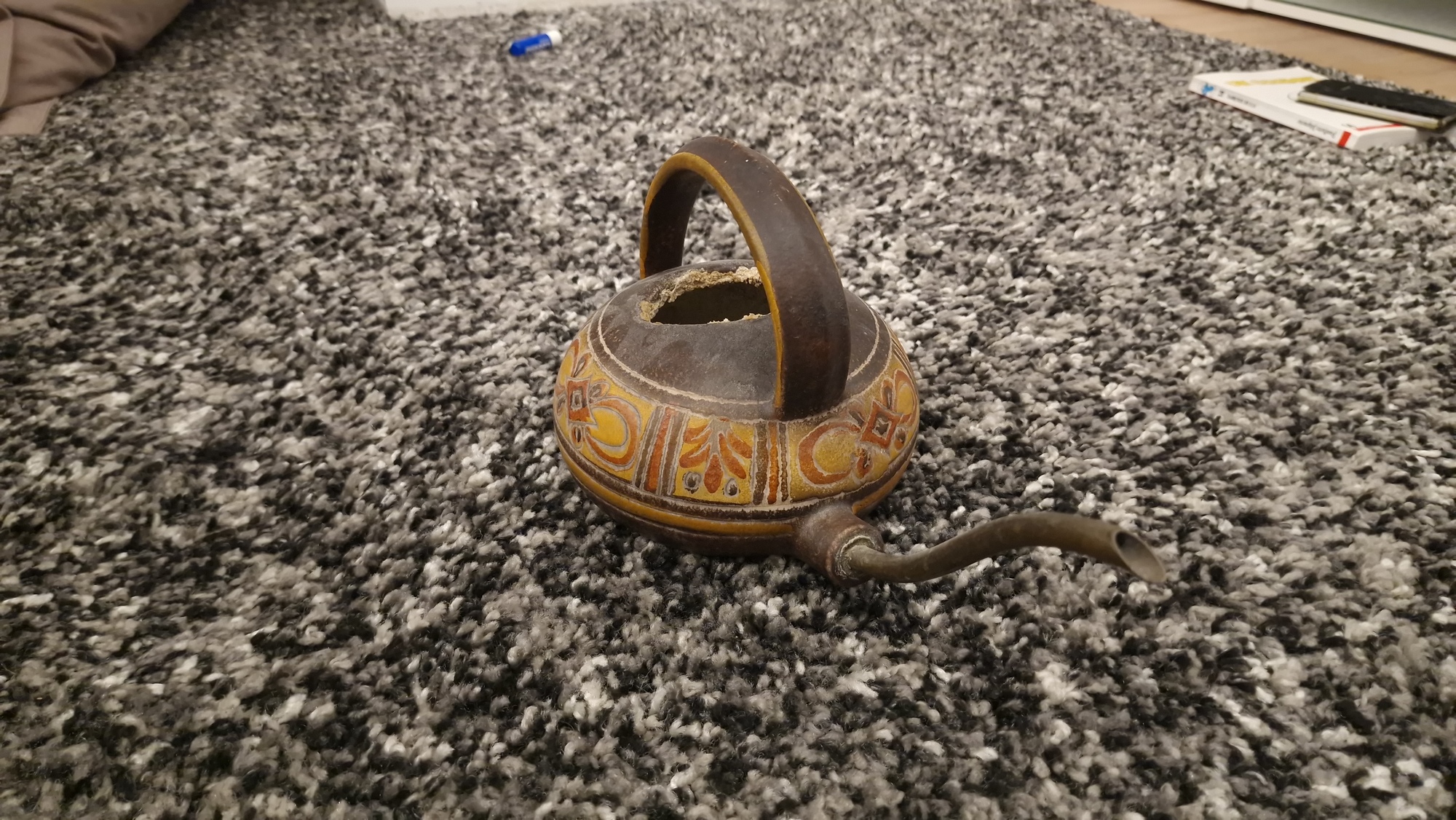}
    &
    \includegraphics[width=0.23\textwidth]{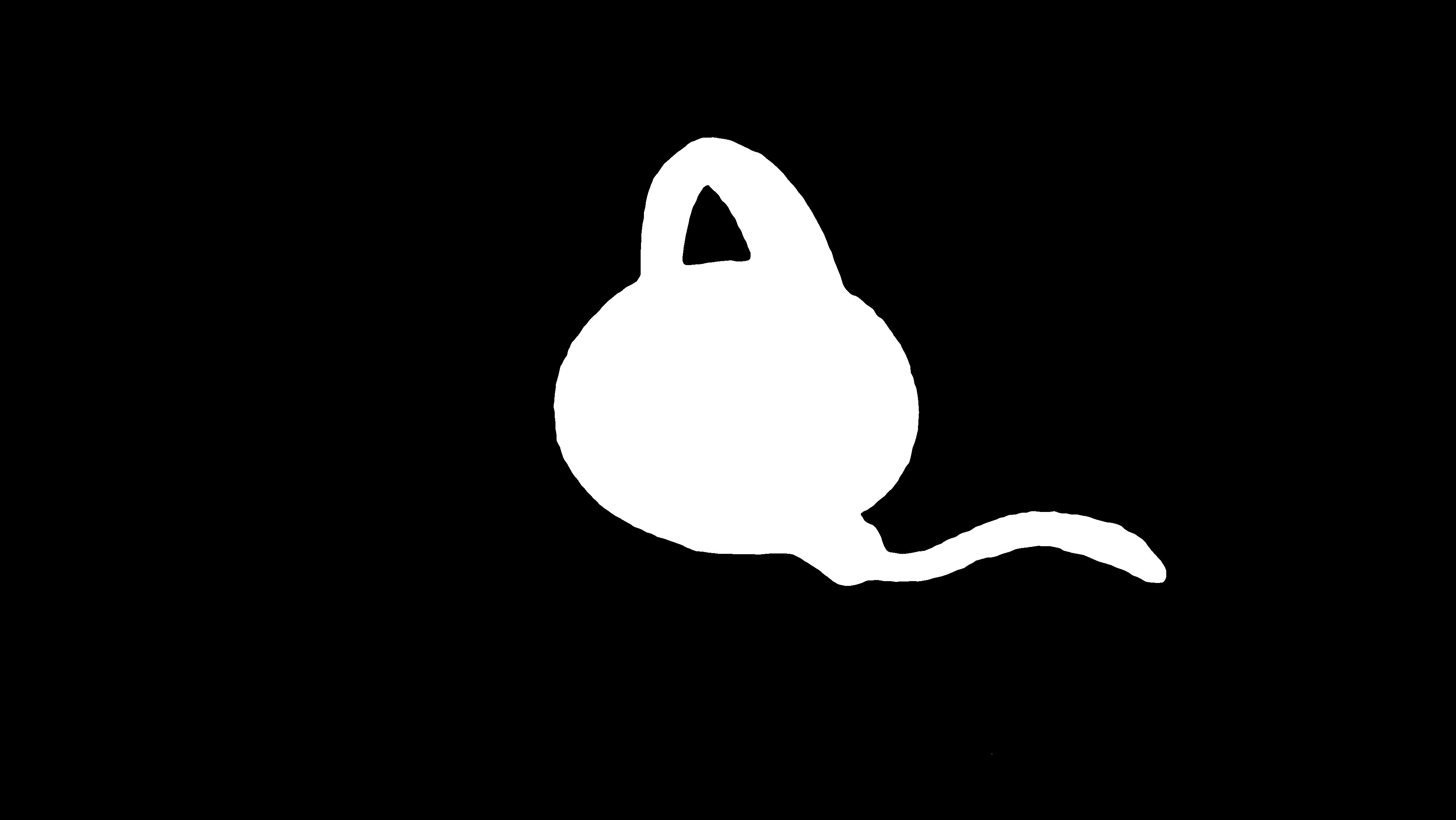}
    \\
    \includegraphics[width=0.23\textwidth]{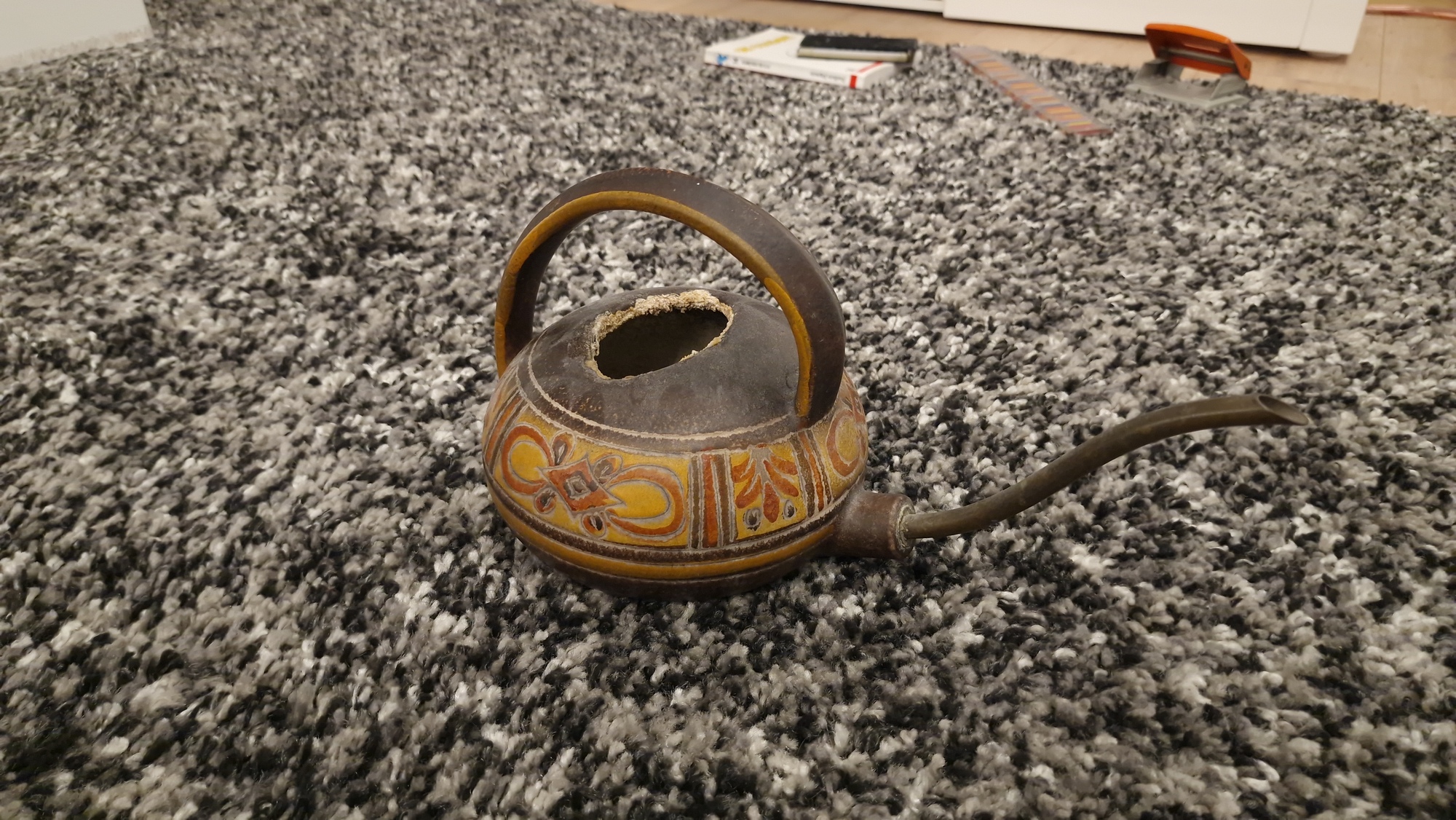}
    &
    \includegraphics[width=0.23\textwidth]{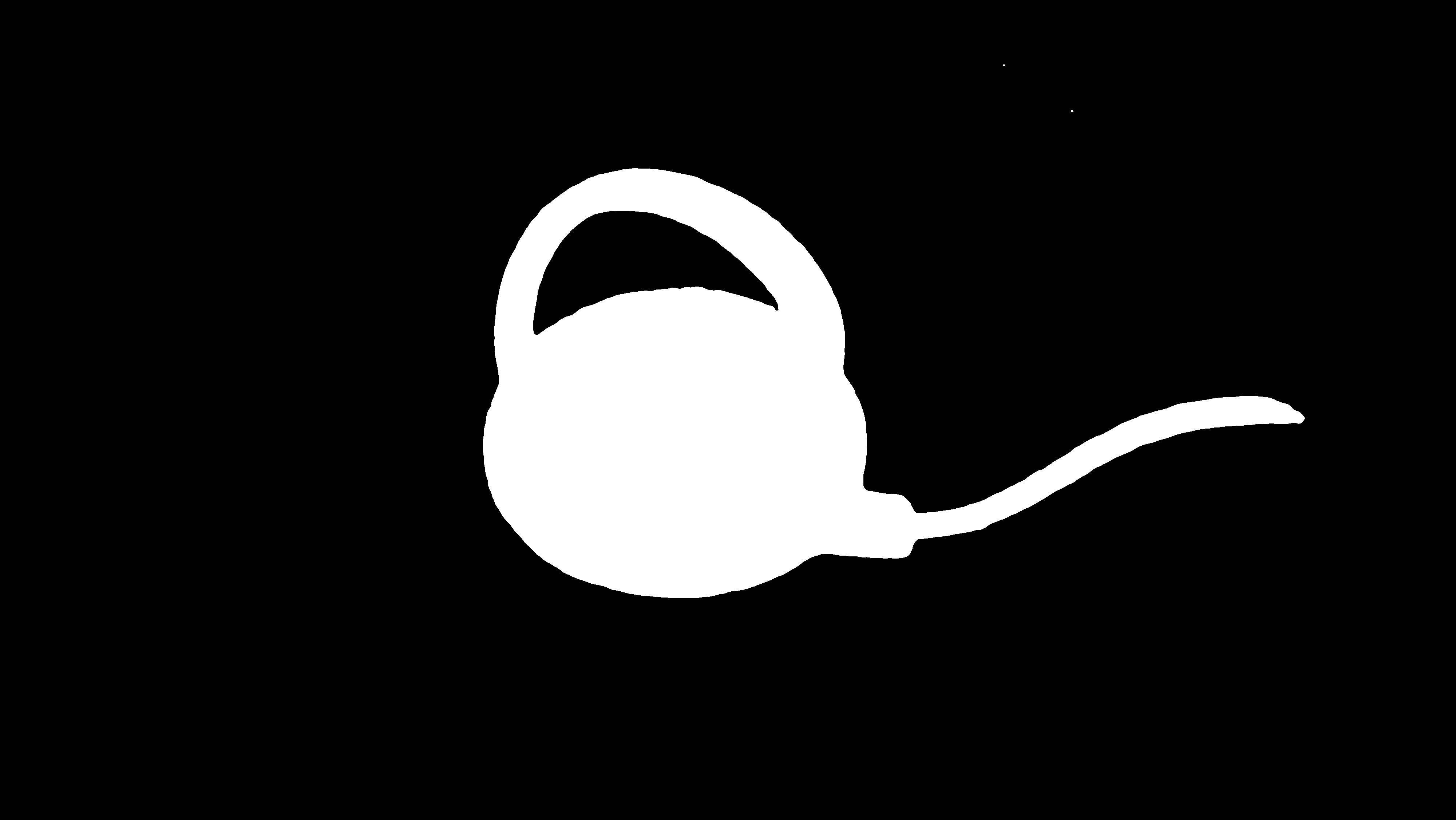}
    \\
    \includegraphics[width=0.23\textwidth]{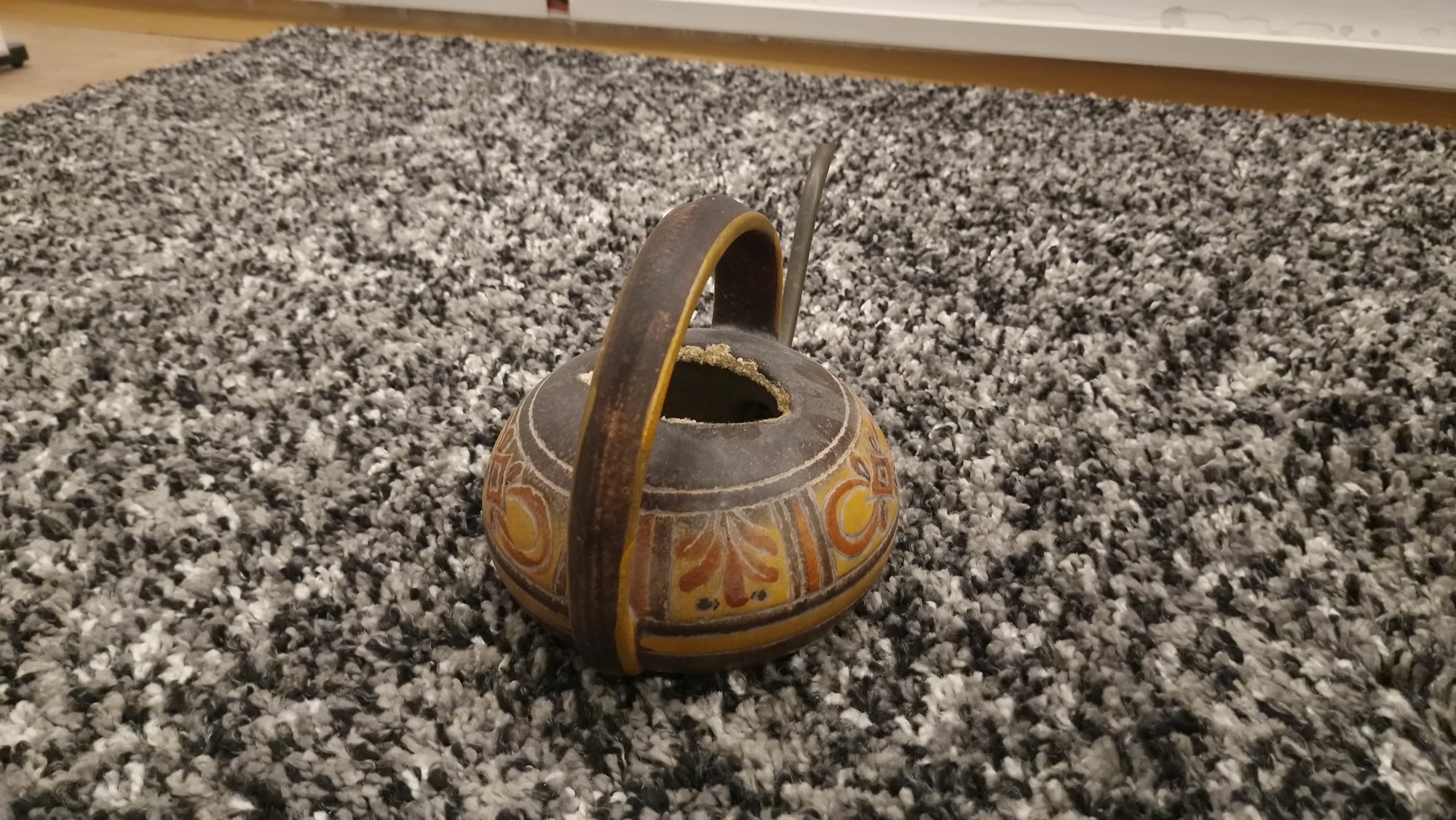}
    &
    \includegraphics[width=0.23\textwidth]{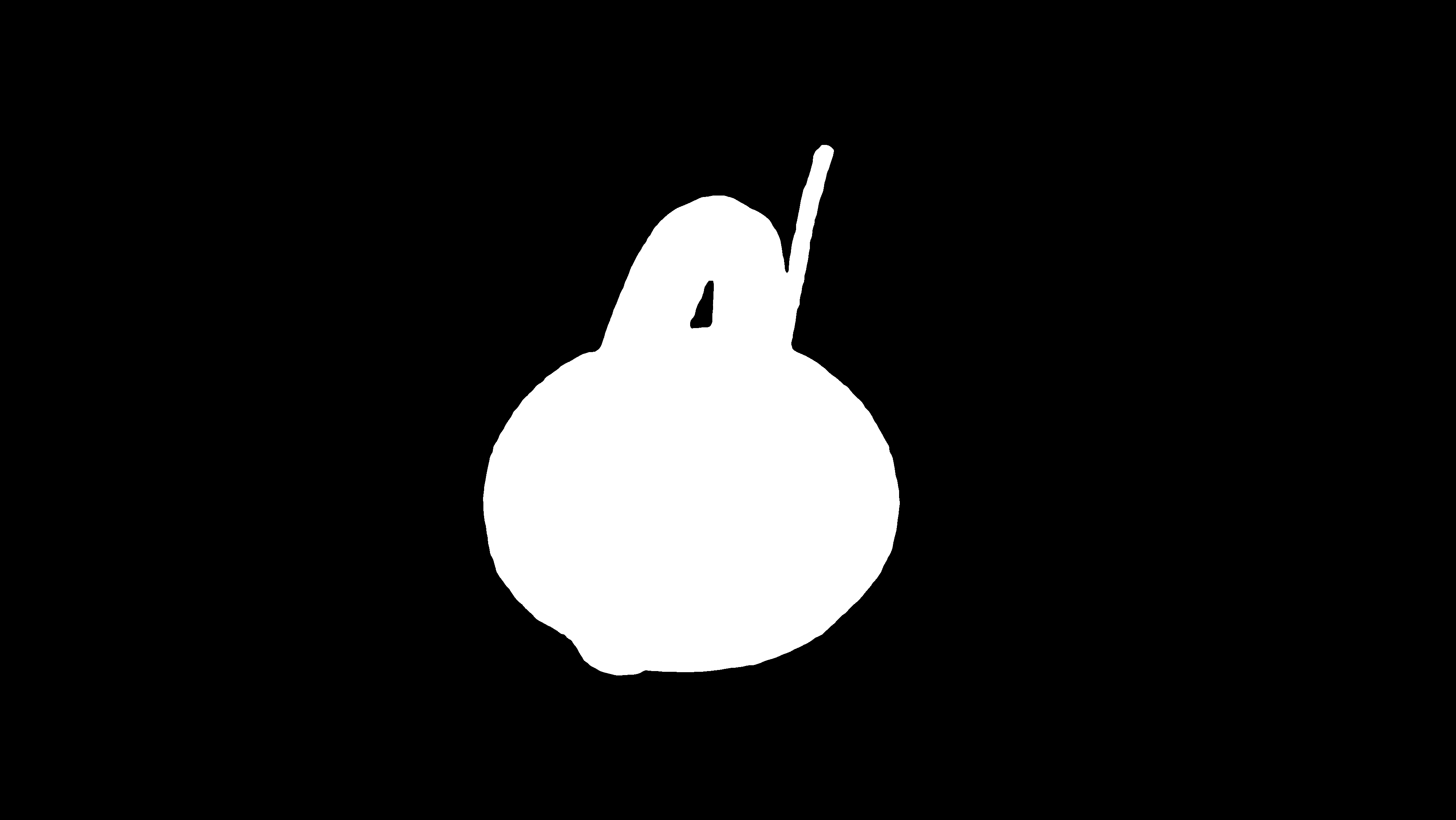}
    \end{tabular}
&
    \setlength\tabcolsep{0pt}
    \renewcommand{\arraystretch}{0}
    \centering
    \begin{tabular}{cc}
    \includegraphics[width=0.23\textwidth]{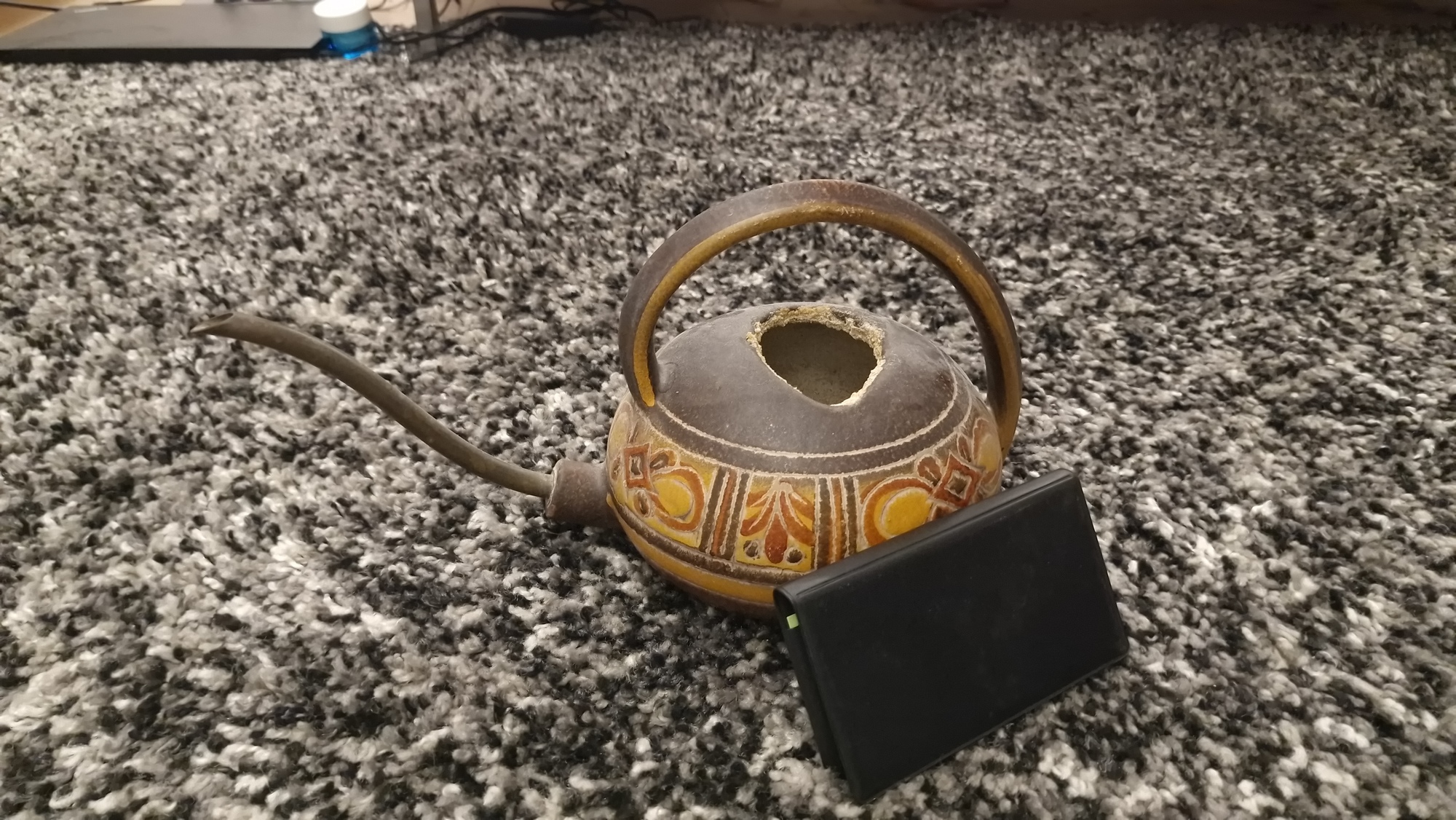}
    &
    \includegraphics[width=0.23\textwidth]{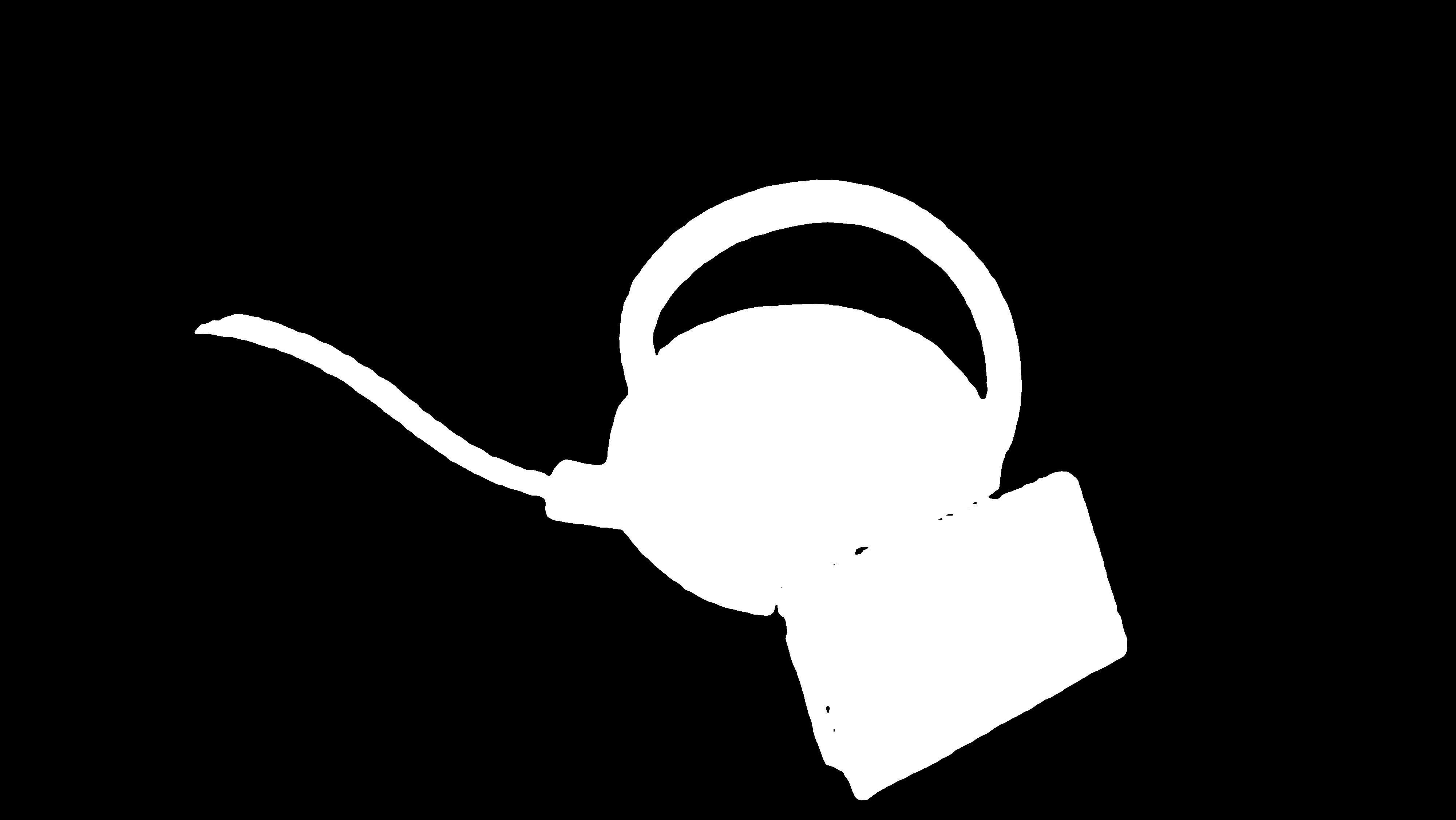}
    \\
    \includegraphics[width=0.23\textwidth]{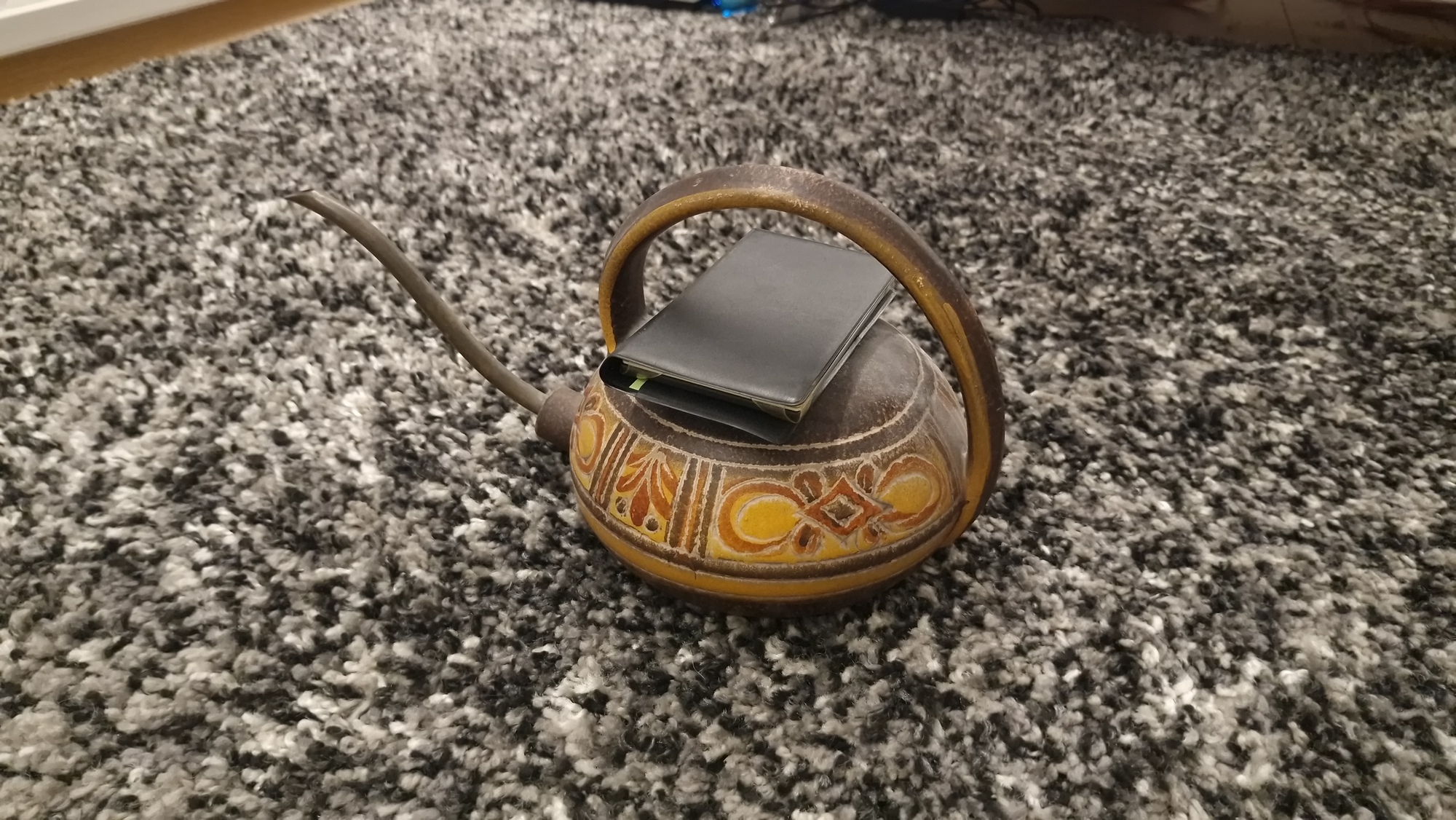}
    &
    \includegraphics[width=0.23\textwidth]{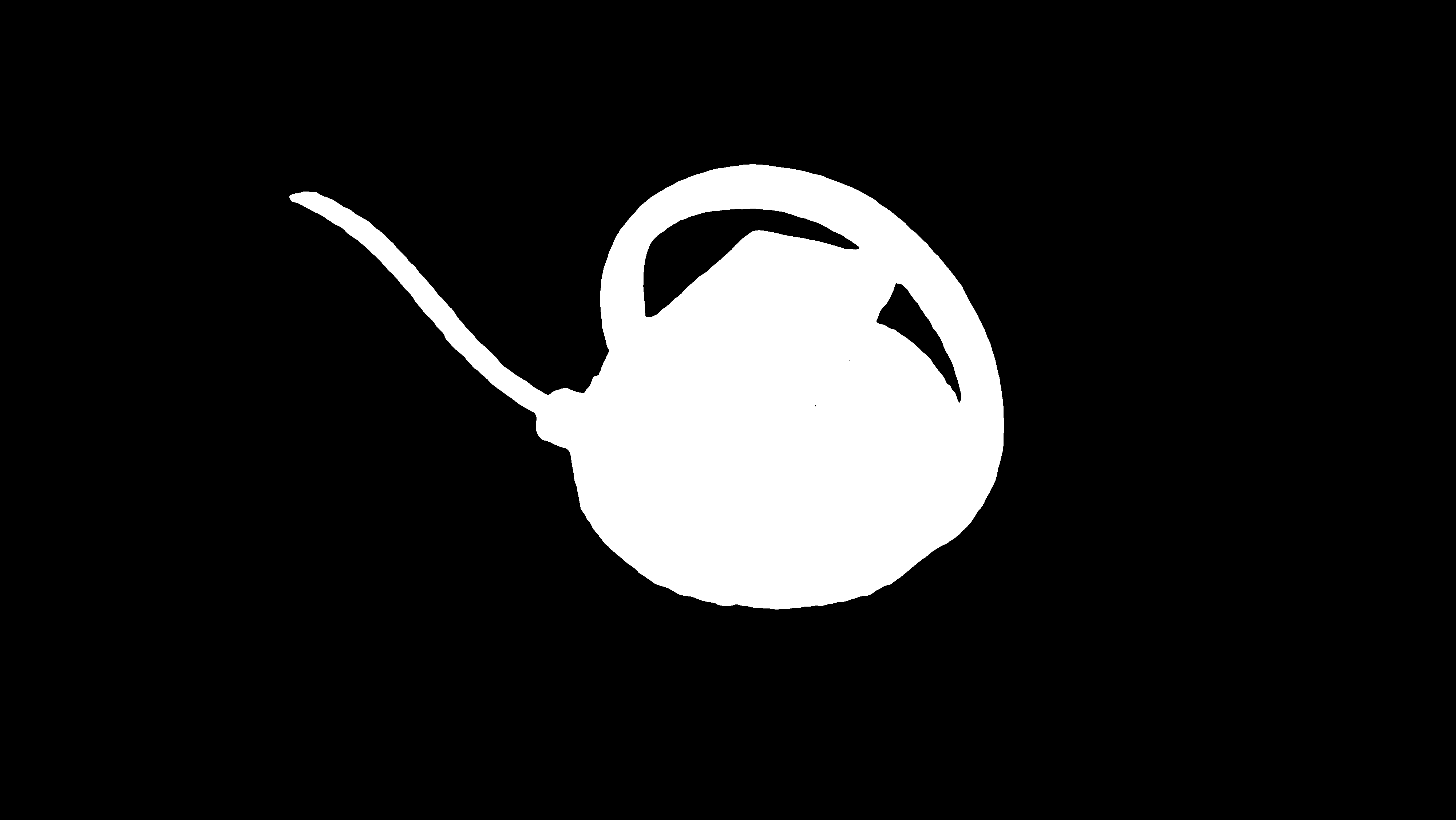}
    \\
    \includegraphics[width=0.23\textwidth]{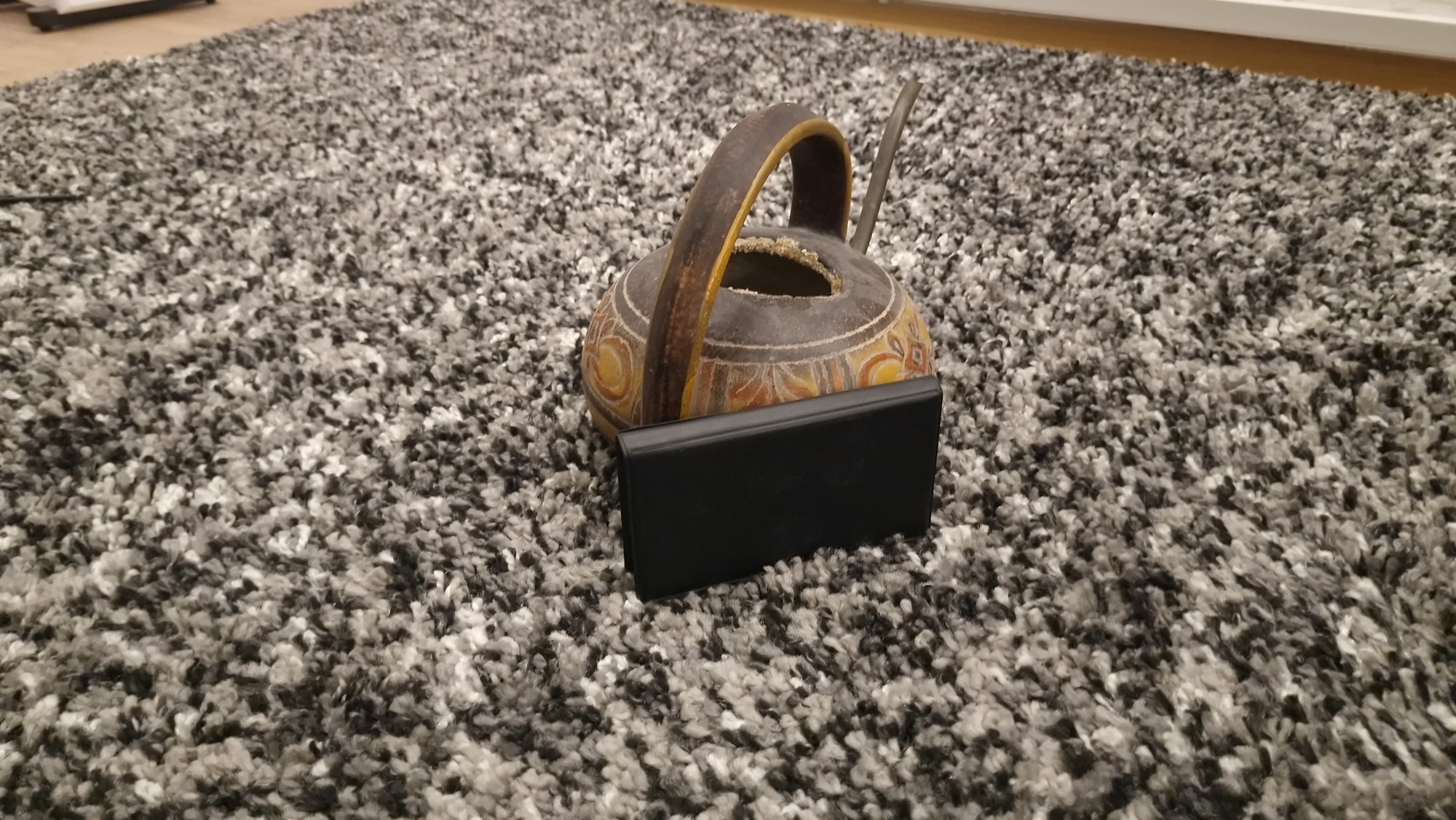}
    &
    \includegraphics[width=0.23\textwidth]{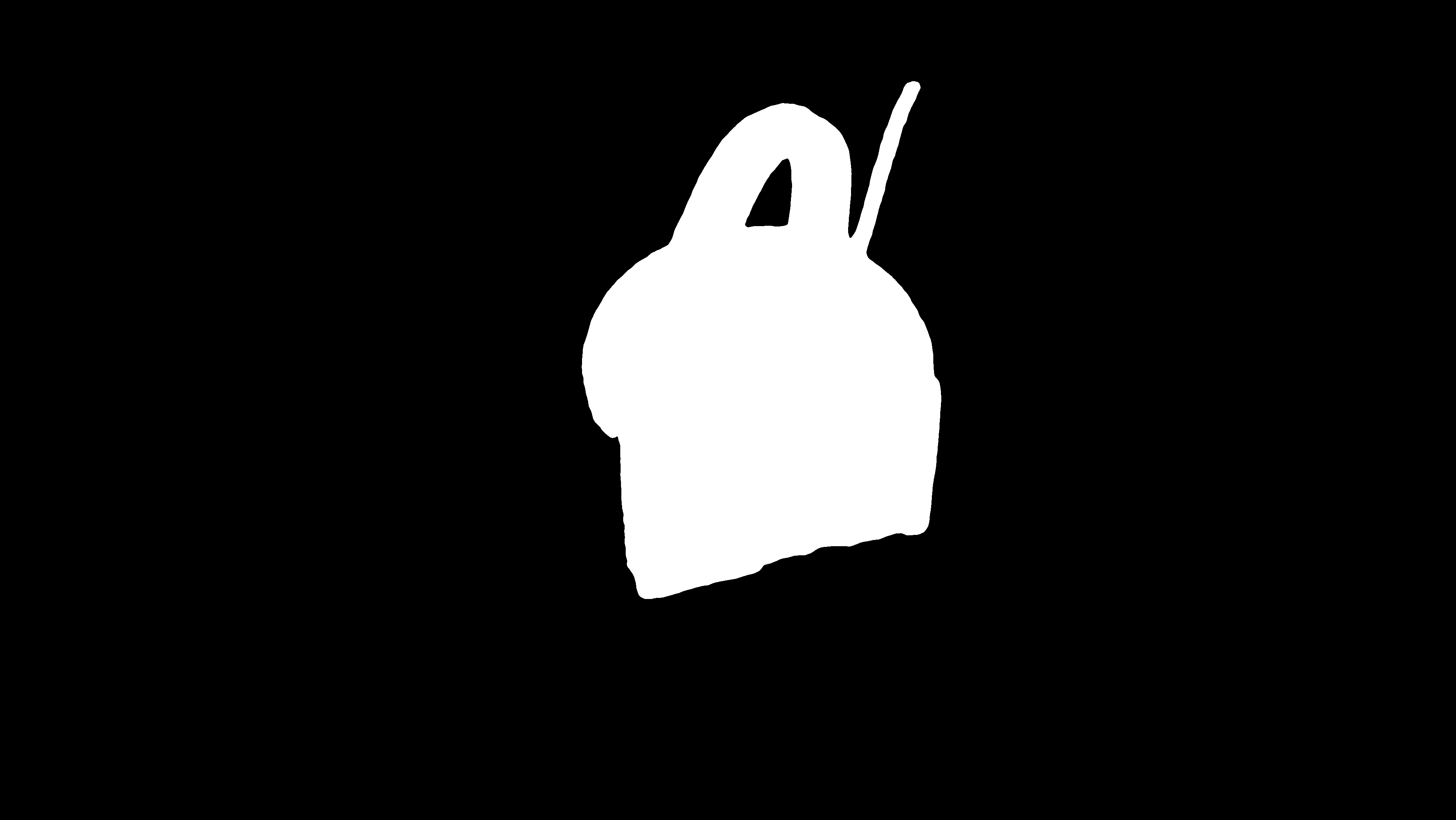}
    \\
    \includegraphics[width=0.23\textwidth]{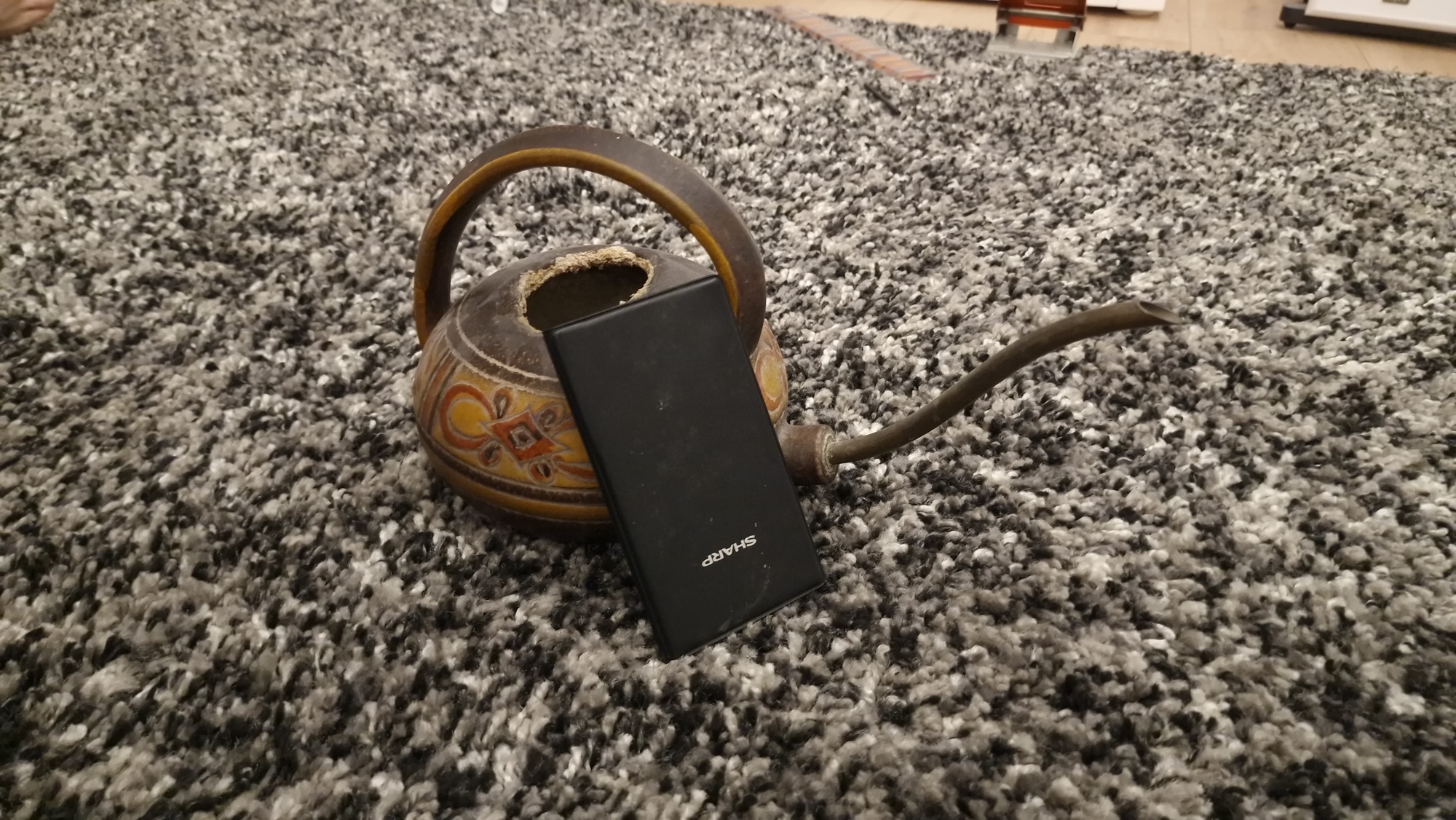}
    &
    \includegraphics[width=0.23\textwidth]{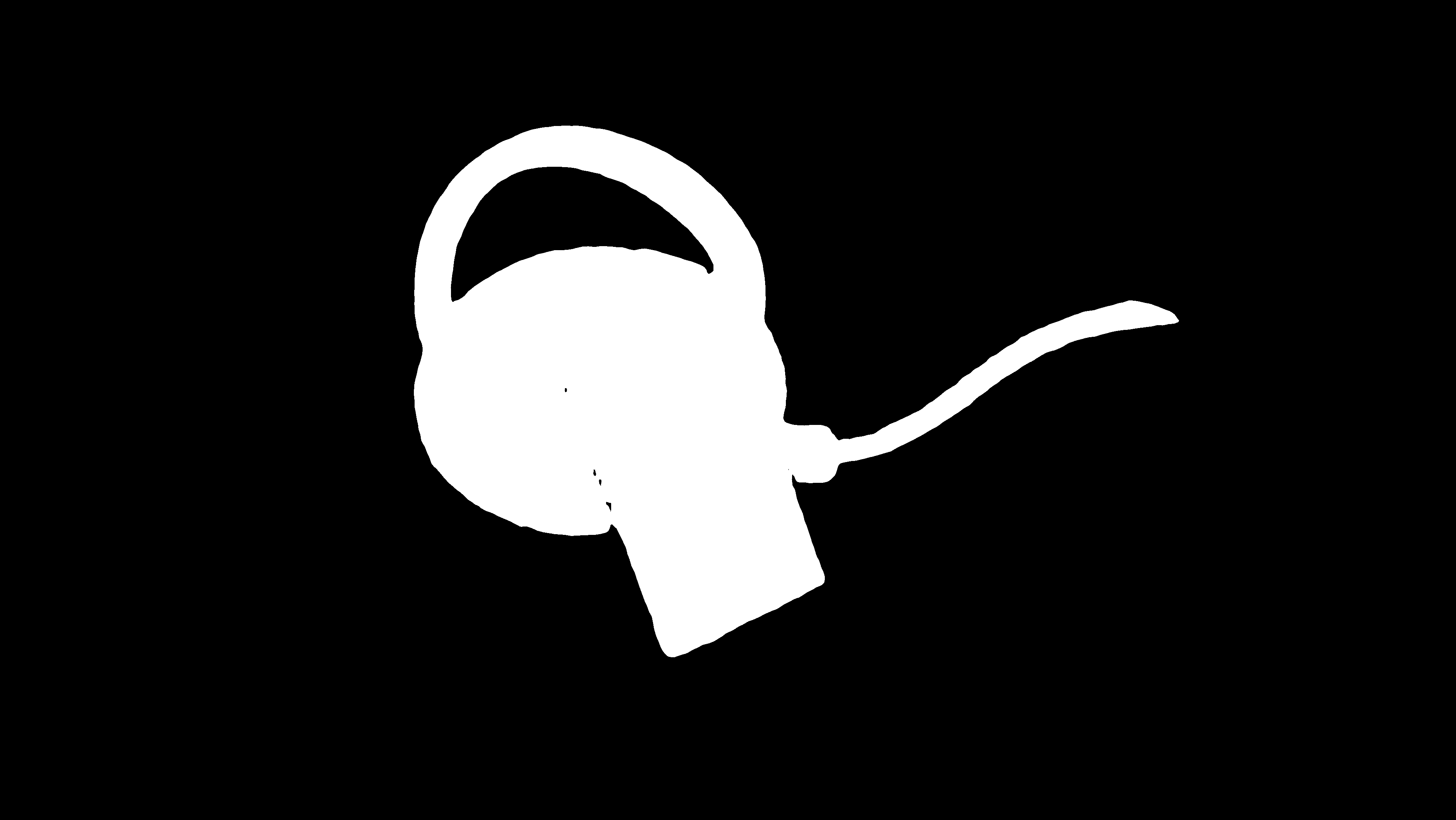}
    \\
    \end{tabular}
\end{tabular}
    \caption{Examples of the clean (left) and occluded (right) images with foreground masks from our Occluded Watering Pot dataset}
    \label{fig:ds_occ}
    \label{fig:ds_clean}
\end{figure}

The watering pot dataset was captured to evaluate the photo-realistic 360-degree reconstruction of a single occluded object from lazily captured real-world images (resembling a typical cultural heritage application). We capture clean and occluded images (Fig. \ref{fig:ds_occ}) from the same range of perspectives using a smartphone camera. The clean images are randomly split into train and test images. The occluded images are partially added to the training images to achieve the desired amount of distraction. Foreground masks containing object and occlusion are automatically obtained using SegmentAnything \cite{kirillov2023segany} with a simple box query. The camera parameters are obtained using COLMAP~\cite{schoenberger2016sfm}. Erroneous estimates of camera parameters or masks are conveniently dealt with by our method, making the entire object reconstruction pipeline robust. The 4000$\times$2252 images are down-scaled by a factor of five before training.

\begin{figure}[!htb]
\includegraphics[width=\textwidth]{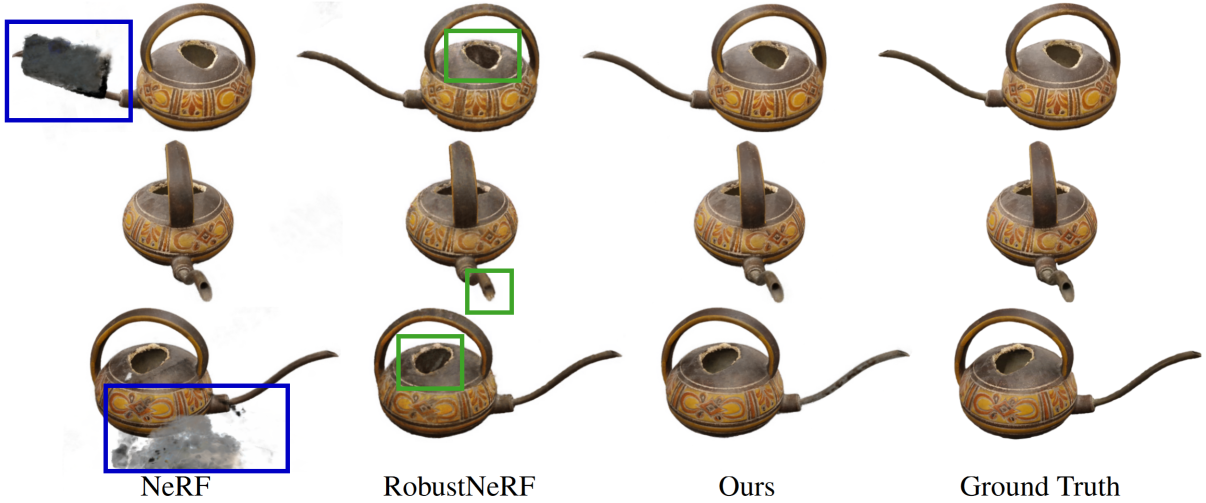}
    \caption{Comparison of reconstruction quality achieved by different methods: Whereas the occlusions lead to well-visible artifacts (blue) in the reconstructions based on NeRF, these artifacts are completely removed by our RANRAC approach. Additionally, the slight inaccuracies of robustNeRF due to the rigorous exclusion of valid samples can be noted (green). They are best visible for view-dependent effects at concavities. Meanwhile, RANRAC preserves these details.}
    \label{fig:nerf_floaters_removed}
\end{figure}

In Figure~\ref{fig:nerf_floaters_removed}, we provide a comparison of the reconstruction quality achieved based on different methods. The artifacts of occluded perspectives are well visible and removed by the robust methods. RobustNeRF's rigorous exclusion of non-photo-consistent samples leads to artifacts for view-dependent details while our RANRAC approach allows a seamless reconstruction even for such details of view-dependent appearance.


\subsection{Inconsistent Camera Parameters}
\begin{figure}
    \centering
    \includegraphics[width=.8\textwidth]{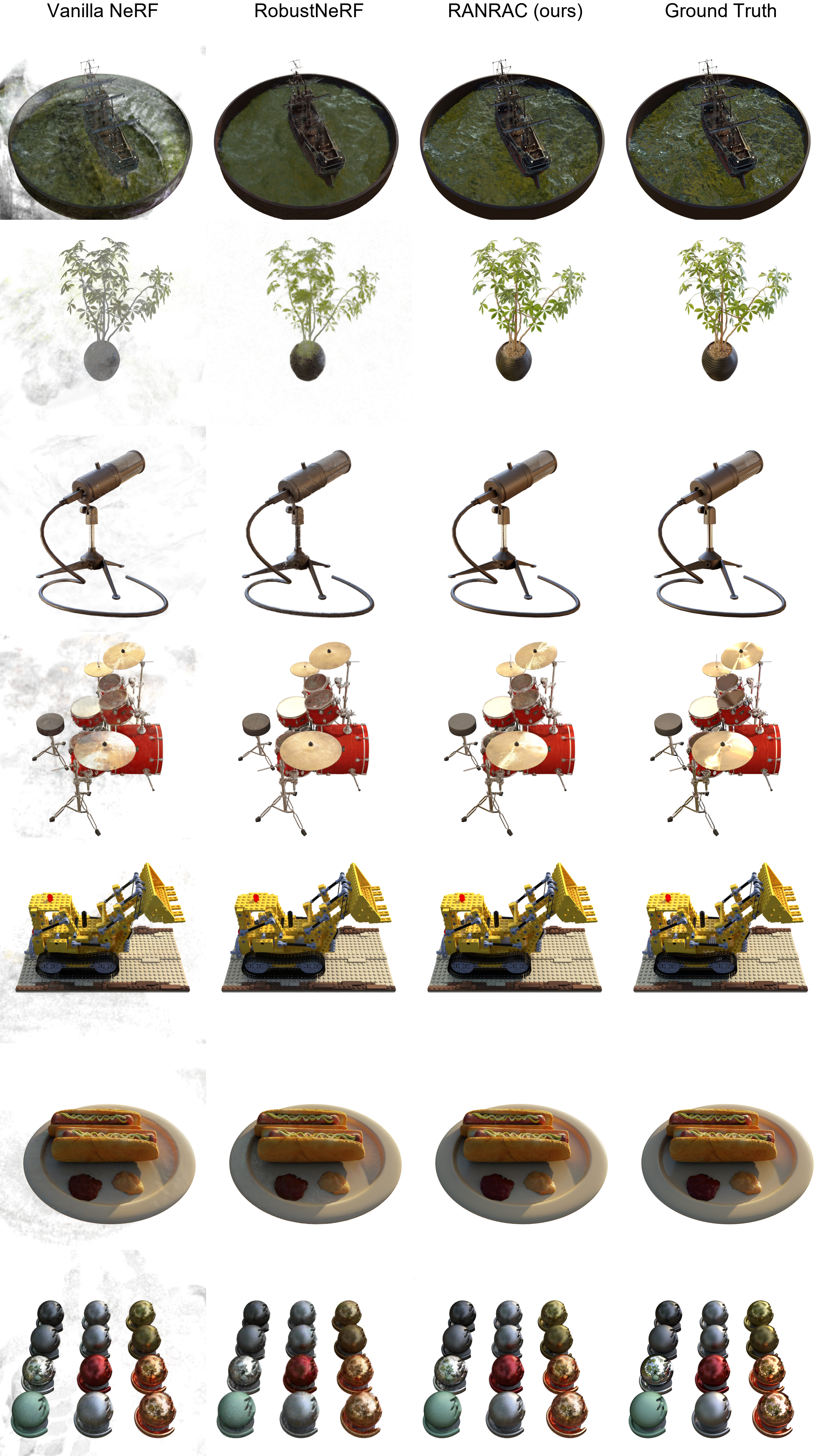}
    \caption{Qualitative results for scenes with noisy camera parameters: The floating artifacts in the NeRF reconstructions are well visible and not present in the reconstructions of RANRAC. Additionally, RobustNeRF rigorously enforces photo-metric consistency, leading to the exclusion of view-dependent specular details (Hi-Hats of drums, specular reflections on the water/microphone/steel balls/flower pot, view-dependent appearance of leaves). RANRAC, on the other hand, preserves these details.}
    \label{fig:nerf_camparam_quali}
\end{figure}

\begin{figure}
    \centering
    \includegraphics[width=.8\textwidth]{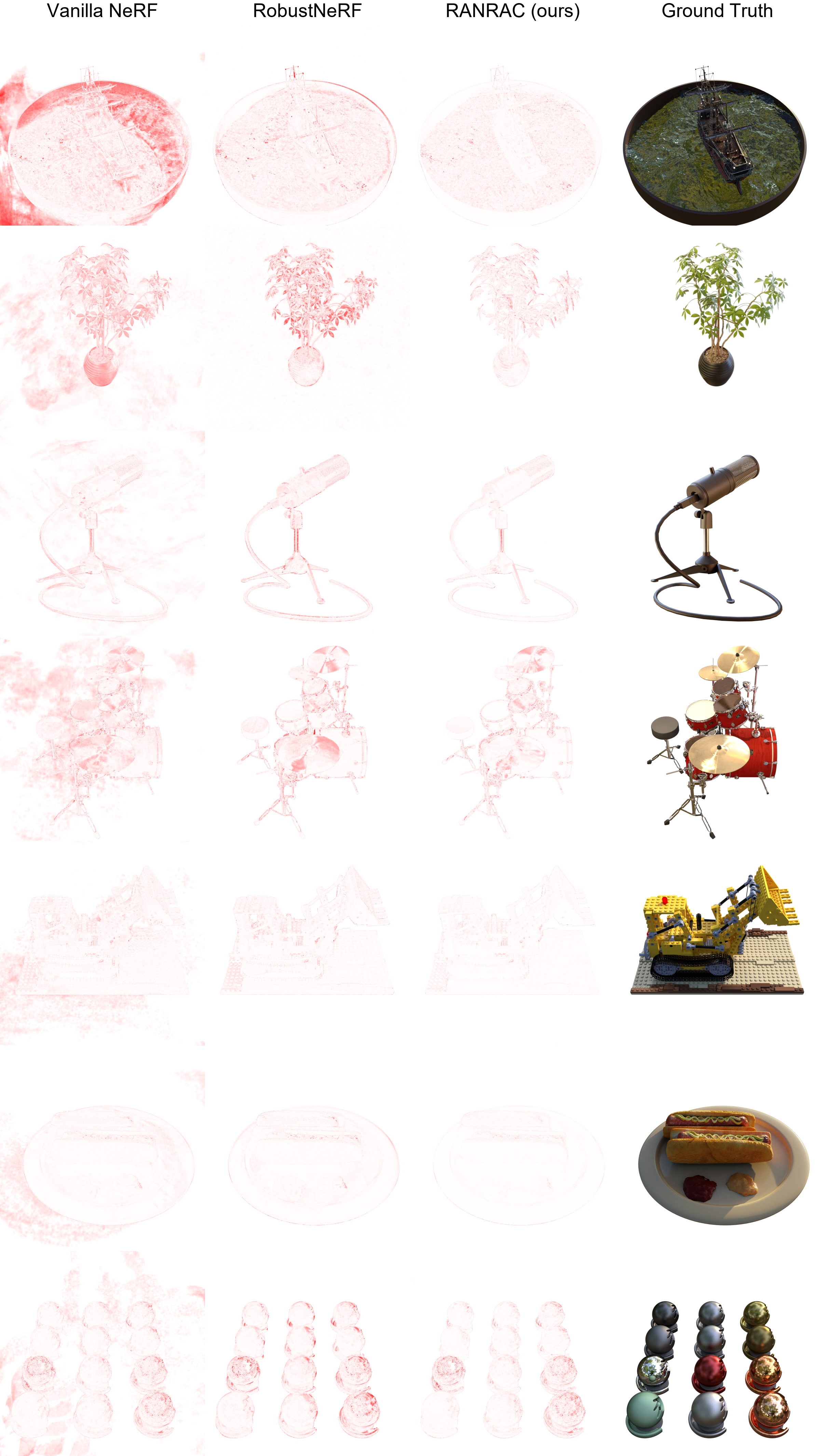}
    \caption{Error visualization for scenes with noisy camera parameters: We observe the global artifacts in the NeRF reconstruction and the local artifacts for scenes with strong view-dependent details in the robustNeRF reconstructions indicated by a reddish highlighting (i.e., the larger the error the better visible it is in the difference image). In contrast, RANRAC does not produce global artifacts and has much lower prediction error for view-dependent details, solely caused by the scene representation itself.}
    \label{fig:nerf_camparam_diff}
\end{figure}

Structure-from-motion pipelines are a powerful tool for pose estimation. It is, however, common that the registration fails for some images. These images will then induce artifacts into the neural field reconstruction. With RANRAC we can simply separate all such perspectives from the training data without removing other valuable information like view-dependent appearance and fit the model to a clean set of observations.
Complementing the evaluation in the main paper, we present additional qualitative results (Fig. \ref{fig:nerf_camparam_quali}) as well as difference images (Fig. \ref{fig:nerf_camparam_diff}) to provide an intuitive assessment regarding how the errors behave across the scene.

\subsection{Unfocused Perspectives}
\begin{figure}
    \centering
    \includegraphics[width=.8\textwidth]{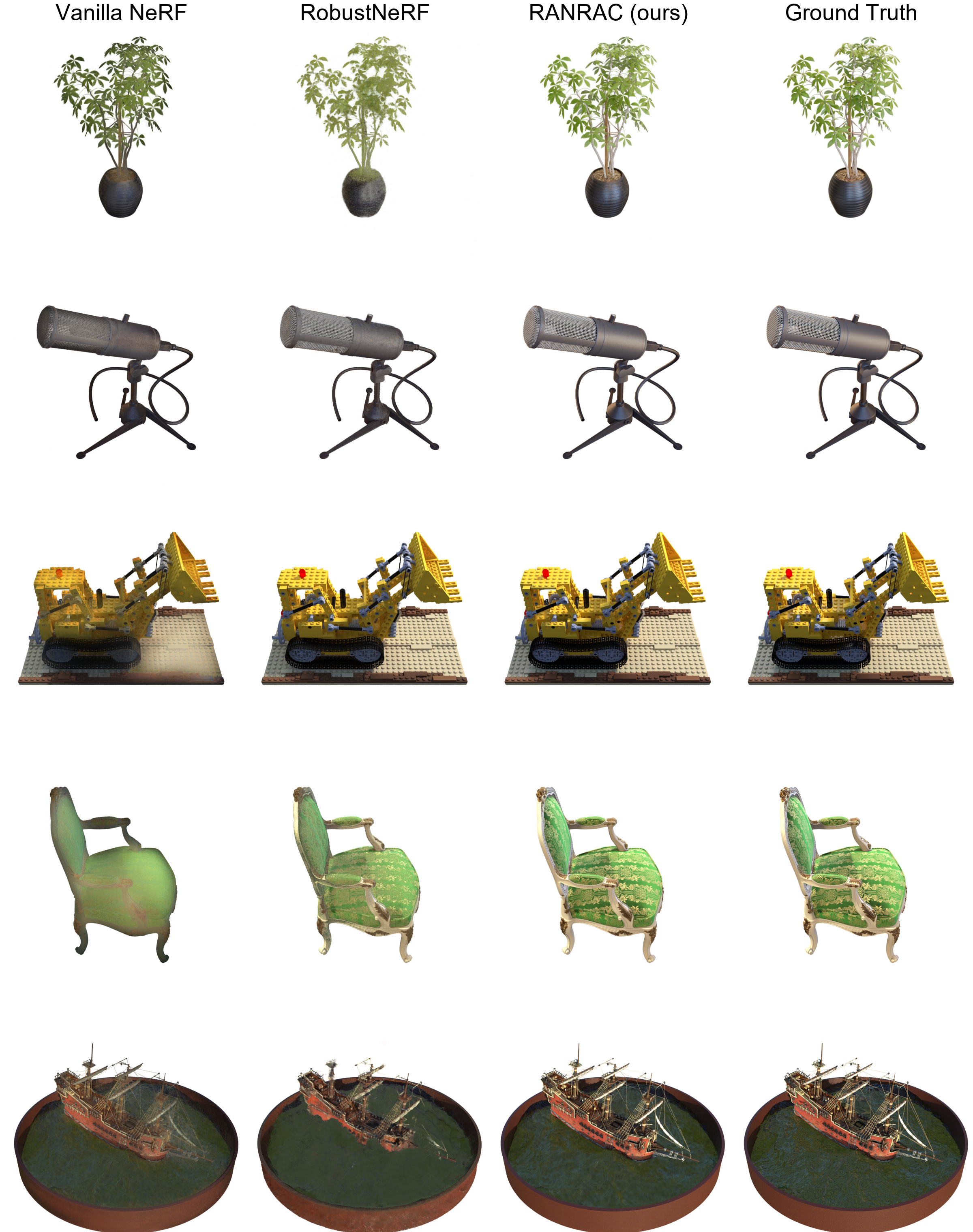}
    \caption{Qualitative results for scenes with unfocused images: RANRAC successfully separates the blurred observations from clean ones while RobustNeRF again struggles with view-dependent appearance.}
    \label{fig:nerf_blur_qual}
\end{figure}

\begin{figure}
    \centering
    \includegraphics[width=.8\textwidth]{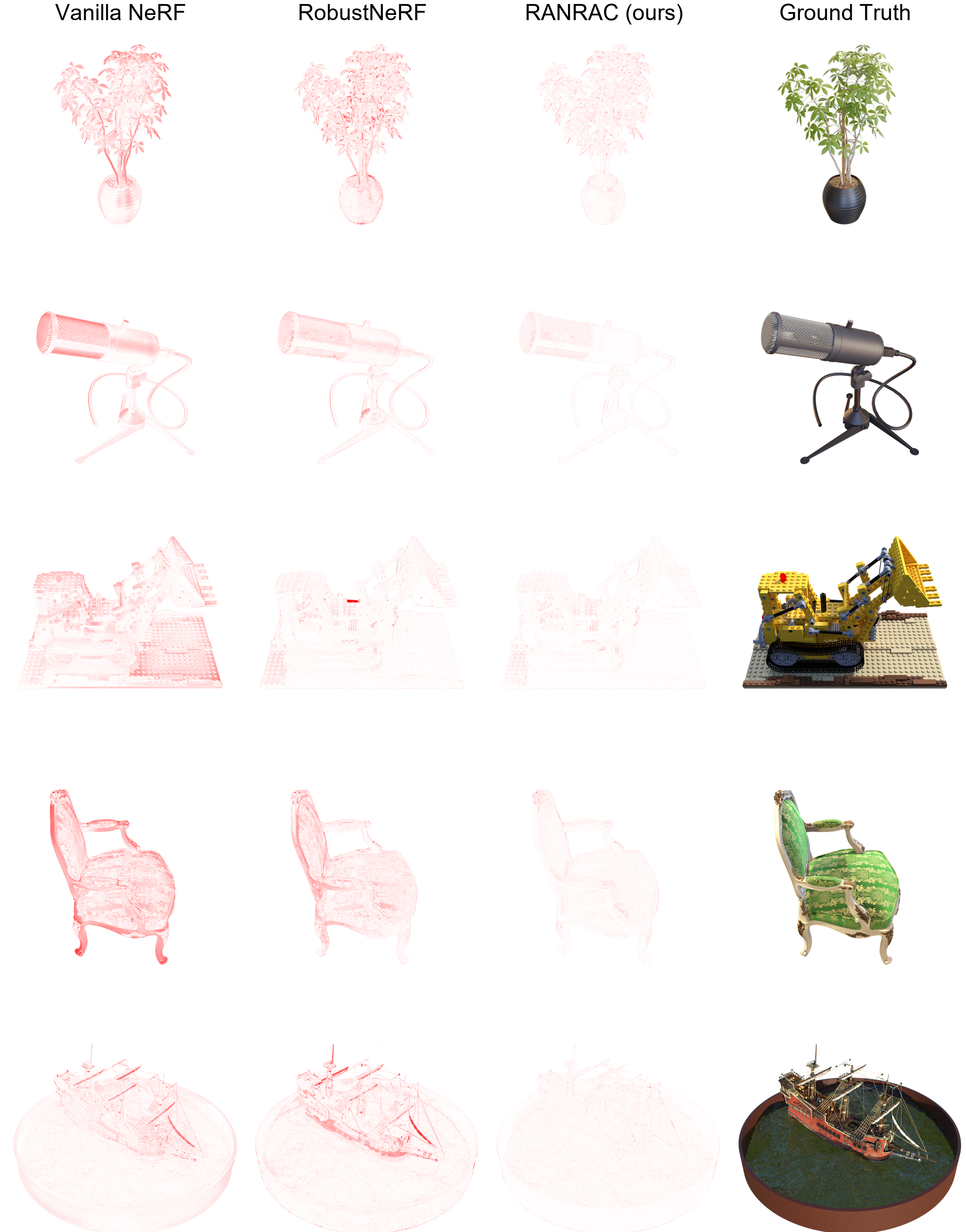}
    \caption{Error visualization for scenes with blurred perspectives: We observe the global artifacts in the NeRF reconstruction and the local artifacts for view-dependent details in the RobustNeRF reconstructions indicated by a reddish highlighting (i.e., the larger the error the better visible it is in the difference image). In contrast, RANRAC does not produce global artifacts and has much lower prediction error for view-dependent details, solely caused by the scene representation itself.}
    \label{fig:nerf_blur_diff}
\end{figure}

\begin{figure}
    \centering
    \includegraphics[width=0.8\textwidth]{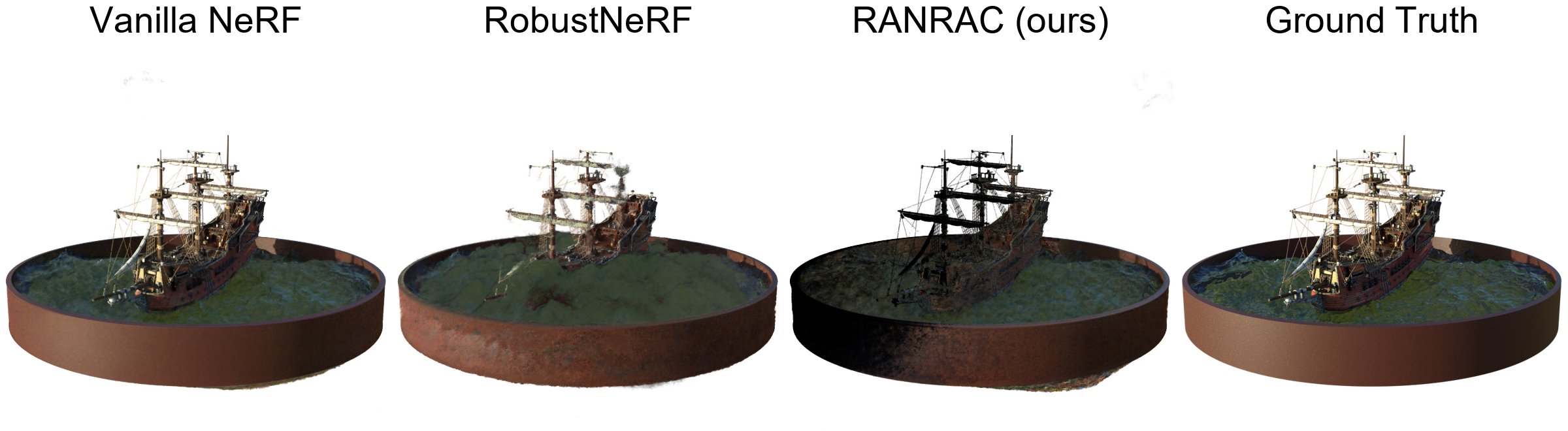}
    \caption{Both robust methods struggle separating the few perspectives of the ship with low viewing angle from the blurred images. While RANRAC still removes blurred perspectives and achieves a strongly improved reconstruction (Figs. \ref{fig:nerf_blur_qual}, \ref{fig:nerf_blur_diff}), for those few perspectives the appearance is not correctly reconstructed.}
    \label{fig:failure_case}
\end{figure}

Within a lazy capture of an object, typically, some of the frames will be out of focus. We also investigate RANRAC's capabilities to deal with such blurred frames in the set of input images. In Figure \ref{fig:nerf_blur_qual}, we complement the results of the main paper with further qualitative results. Additionally, we present difference images in Figure \ref{fig:nerf_blur_diff} to better illustrate where errors are larger or smaller respectively. For sparse and irregular viewing angles both robust methods struggle separating them from blur resulting in RANRAC's only failure case (Fig. \ref{fig:failure_case}). This could be addressed with additional priors.

\clearpage
\bibliographystyle{splncs04}
\bibliography{egbib}

%% file: figures/teaser.tex
\begin{figure}
    \centering
    \includegraphics[width=0.7\linewidth]{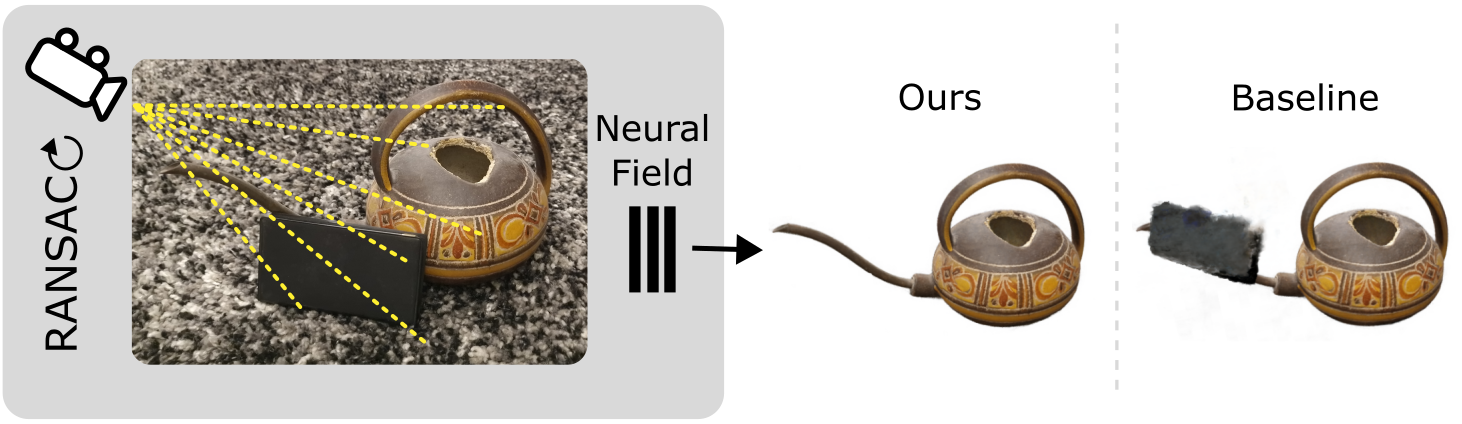}
    \caption{We propose a robust algorithm for 3D reconstruction from occluded input perspectives that is based on the random sampling of hypotheses. Our algorithm is general and we demonstrate the use for single-shot reconstruction using light field networks or multi-view reconstruction using NeRF. In these cases, it successfully removes the artifacts that normally occur due to occluded input perspectives.}
    \label{fig:teaser}
\end{figure}

%% file: figures/method.tex
\begin{figure*}[t]
    \centering
    \includegraphics[width=\linewidth]{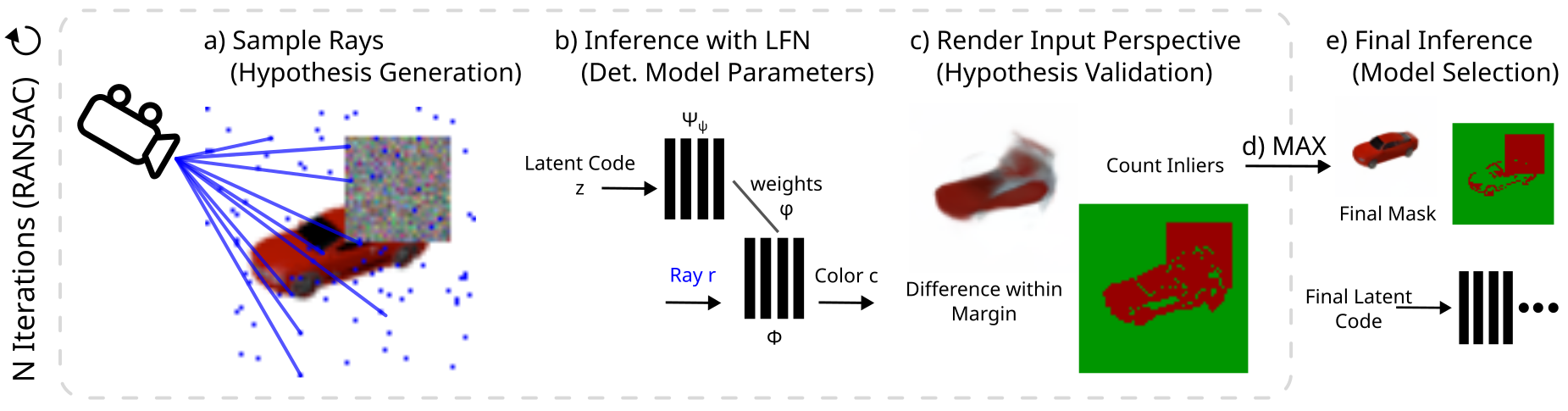}
    \caption{
    The RANRAC algorithm for LFNs samples random hypotheses by selecting a set of random samples from the given perspective (a), and inferring the latent representation of these rays using the autodecoder of a pretrained LFN (b).
    The obtained light field is then used to predict an image from the input perspective (c).
    Based on this prediction, confidence in the random hypothesis is evaluated via the Euclidean distance between the predicted ray colors and the remaining color samples of the input image. The amount of samples which are explained by each hypothesis up to some margin are used to determine the best hypothesis (d).
    All samples explained by the selected hypothesis are used for a final inference with the LFN to obtain the final model and latent representation (e).
    }
    \label{fig:ranrac_overview}
\end{figure*}

%% file: figures/lfn_table.tex
\begin{table}
\vspace{-0.5cm}
    \centering
    \scriptsize
\setlength\tabcolsep{2.5pt}

\begin{tabular}{c|c||cccccccccc}
Metric & Model & Bench & Boat & Cabin. & Displ. & Lamp & Phone & Rifle & Sofa & Speak. & Table
\\ \hhline{=|=#==========}
\multirow{2}{*}{PSNR$\uparrow$} &
RANRAC & \bfseries 19.21 & \bfseries 22.92 & \bfseries 21.92 & 17.65 & \bfseries 19.99 & \bfseries 18.45 & \bfseries 21.35 & \bfseries 21.61 & \bfseries 20.7 & \bfseries 20.44
\\
{} & LFN & 17.89 & 19.2 & 20.5 & \bfseries 18.85 & 19.09 & 17.81 & 18.46 & 20.12 & 19.99 & 20.29
\\

\multirow{2}{*}{SSIM$\uparrow$} &
RANRAC & \bfseries 0.767 & \bfseries 0.858 & \bfseries 0.801 & 0.699 & \bfseries 0.764 & \bfseries 0.75 & \bfseries 0.853 & \bfseries 0.805 & \bfseries 0.761 & \bfseries 0.784
\\
{} & LFN & 0.724 & 0.791 & 0.767 & \bfseries 0.73 & 0.748 & 0.726 & 0.795 & 0.775 & 0.743 & 0.777
\\

\end{tabular}
\caption{RANRAC obtains a significant quantitative improvement in PSNR and SSIM (higher is better) compared to the baseline. We compare RANRAC to vanilla LFNs for the 13 largest ShapeNet classes (find plane, car, and chair with more detail in Fig.~\ref{fig:recransacoccimg}). The results are based on a moderate amount of occlusion of approximately 25\% object occlusion and about 5\% image occlusion. The reported results are conservative, as higher amounts of image occlusion result in a more significant performance increase (Fig. \ref{fig:recransacoccimg}). No  hyperparameter tuning has been performed for these classes; the same configuration obtained from the analysis of the other three classes is used.}
\vspace{-1cm}
    \label{tab:quant_other_classes}
\end{table}

%% file: figures/lfn_plot_imgocc.tex
\begin{figure}
\begin{tikzpicture}
    \begin{axis}[
        width=.4\textwidth,
        height=4cm,
        xlabel={Image Occlusion in \%},
        ylabel={PSNR},
        y label style={at={(-0.12,0.8)}},
        name=psnr,
        xmajorgrids,
        mark size=1pt,
        cycle list name = foo bar,
        legend style={
            at={(1.0,-0.5)},
            anchor=north,
            nodes={scale=0.7, transform shape},
            draw=none},
        legend columns=6,
        label style={font=\scriptsize},
        tick label style={font=\scriptsize},
        separate axis lines,
        axis y line*=left,
        axis x line*=bottom,
        y axis line style= { draw opacity=0 },
        x axis line style= { draw opacity=0 },
        xmin=0,
        xmax=45,
        ymin=11,
        ymax=30,
        ]
    
    \addplot coordinates {
(0, 26.680033408865164)(5, 25.571872120289402)(15, 20.723016363764444)(25, 18.452052133857276)(35, 14.05213147492478)(45, 12.598868768047424)
        };
    \addlegendentry{Plane Ours}

    \addplot coordinates {
(0, 28.95809361439043)(5, 20.841208262771993)(15, 15.263018360063226)(25, 13.496575865310241)(35, 12.192475953847111)(45, 11.652292588514124)
        };
    \addlegendentry{Plane LFN}

    \addplot coordinates {
(0, 25.633840034259354)(5, 24.450873637293828)(15, 23.207895082244654)(25, 21.5986293898245)(35, 18.339842088145087)(45, 13.134919200871877)
        };
    \addlegendentry{Car Ours}

    \addplot coordinates {
(0, 27.5457678032748)(5, 22.434692475892593)(15, 14.929294975676633)(25, 13.111142323337434)(35, 12.605926332480557)(45, 11.519278067747608)
        };
    \addlegendentry{Car LFN}

    \addplot coordinates {
(0, 20.16970127461686)(5, 19.405604581302786)(15, 17.059901259568726)(25, 16.06942908372162)(35, 15.016639277048004)(45, 13.148959144858102)
        };
    \addlegendentry{Chair Ours}

    \addplot coordinates {
(0, 21.79202911544893)(5, 19.866299275191064)(15, 14.829935844979355)(25, 13.49238624668675)(35, 12.679118062185196)(45, 11.778441171498875)
        };
    \addlegendentry{Chair LFN}

    \legend{};
    \end{axis}
    \hspace{.7cm}
    \begin{axis}[
        width=.4\textwidth,
        height=4cm,
        xlabel={Image Occlusion in \%},
        ylabel={SSIM},
        y label style={at={(-0.14,0.8)}},
        legend style={
            at={(0.5,-0.55)},
            anchor=south,
            nodes={scale=0.7, transform shape},
            draw=none,
            },
        legend columns=6,
        name=ssim,
        at=(psnr.north east),
        anchor=north west,
        xmajorgrids,
        mark size=1pt,
        cycle list name = foo bar,
        xlabel near ticks,
        label style={font=\scriptsize},
        tick label style={font=\scriptsize},
        separate axis lines,
        axis y line*=left,
        axis x line*=bottom,
        y axis line style= { draw opacity=0 },
        x axis line style= { draw opacity=0 },
        xmin=0,
        xmax=45,
        ymin=0.4,
        ymax=0.95,
        ]
    
    \addplot coordinates {
(0, 0.9116395835557397)(5, 0.8992860307286984)(15, 0.7811481210617672)(25, 0.7107215897853032)(35, 0.5877294073829796)(45, 0.5428376551026892)
        };
    \addlegendentry{Plane Ours}

    \addplot coordinates {
(0, 0.9276764412622966)(5, 0.8211082922056401)(15, 0.6213971914009379)(25, 0.533983198742538)(35, 0.5098133177222111)(45, 0.48294348682611343)
        };
    \addlegendentry{Plane LFN}

    \addplot coordinates {
(0, 0.9220206870960937)(5, 0.9073303904740332)(15, 0.8853276051854598)(25, 0.8371109852861395)(35, 0.7445750207457332)(45, 0.5477536292382214)
        };
    \addlegendentry{Car Ours}

    \addplot coordinates {
(0, 0.9359108380666032)(5, 0.8795253470580165)(15, 0.6318766311673953)(25, 0.5493371159103877)(35, 0.5195514488137251)(45, 0.4549606725913515)
        };
    \addlegendentry{Car LFN}

    \addplot coordinates {
(0, 0.7654769199232196)(5, 0.7419467787015448)(15, 0.6688738807598029)(25, 0.6301808922186047)(35, 0.5896063074139322)(45, 0.4969113391214927)
        };
    \addlegendentry{Chair Ours}

    \addplot coordinates {
(0, 0.7986534410502112)(5, 0.7526394556265085)(15, 0.5697925272394095)(25, 0.4952181717059984)(35, 0.4633653098831412)(45, 0.42372534157515673)
        };
    \addlegendentry{Chair LFN}

    
    \end{axis}
    \hspace{0.7cm}
    \begin{axis}[
        width=.4\textwidth,
        height=4cm,
        xlabel={Object Occlusion in \%},
        ylabel={PSNR},
        y label style={at={(-0.12,0.8)}},
        legend pos=outer north east,
        name=psnrobj,
        at=(ssim.north east), anchor=north west,
        xmajorgrids,
        mark size=1pt,
        cycle list name = foo bar,
        xlabel near ticks,
        legend style={
            nodes={scale=0.7, transform shape},
            draw=none},
        legend columns=1,
        label style={font=\scriptsize},
        tick label style={font=\scriptsize},
        axis y line*=left,
        axis x line*=bottom,
        y axis line style= { draw opacity=0 },
        x axis line style= { draw opacity=0 },
        xmin=-1,
        xmax=76,
        ymin=13,
        ymax=30,
        ]
    \addplot coordinates {
(0, 26.680033408865164)(5, 26.287680269192453)(15, 25.853554304365204)(25, 25.59219147092408)(35, 25.097757073347122)(45, 24.192147399794216)(55, 22.626514970988914)(65, 19.536179828438012)(75, 17.470407216121647)
        };
    \addlegendentry{Plane Ours}

    \addplot coordinates {
(0, 28.95809361439043)(5, 21.009703034098084)(15, 20.79327523825465)(25, 21.396266021951163)(35, 21.89447515909958)(45, 20.963028252703058)(55, 19.397262606068313)(65, 17.55393780722501)(75, 16.23627580074041)
        };
    \addlegendentry{Plane LFN}

    \addplot coordinates {
(0, 25.633840034259354)(5, 25.04387048919001)(15, 24.816514249875155)(25, 24.26340766822573)(35, 23.782224305039627)(45, 22.96312855758541)(55, 21.604715120168493)(65, 20.17751673621323)(75, 16.934796579000658)
        };
    \addlegendentry{Car Ours}

    \addplot coordinates {
(0, 27.5457678032748)(5, 20.82722381654461)(15, 22.014153436974443)(25, 22.03105787720997)(35, 21.929241646256106)(45, 21.054628236564376)(55, 19.053609812947556)(65, 16.901351189623494)(75, 15.215699734988)
        };
    \addlegendentry{Car LFN}

    \addplot coordinates {
(0, 20.16970127461686)(5, 19.450725421369043)(15, 19.23582922564043)(25, 18.312324877784235)(35, 17.986746141098635)(45, 17.74440648792118)(55, 16.59719063333274)(65, 15.689779275988908)(75, 14.96405083300862)
        };
    \addlegendentry{Chair Ours}

    \addplot coordinates {
(0, 21.79202911544893)(5, 18.789964323672145)(15, 19.035795628989714)(25, 18.104309552574495)(35, 17.534599565697203)(45, 17.054960575790197)(55, 15.876874769757576)(65, 15.01713900528394)(75, 13.754751203810416)
        };
    \addlegendentry{Chair LFN}

    \legend{};
    
    \end{axis}
\end{tikzpicture}
\caption{
RANRAC (solid lines) leads to a quantitative improvement in PSNR and SSIM (higher is better) for occluded inputs compared to vanilla LFNs (dashed lines). The same hyperparameter configuration and LFN is used for all classes.
On the left and in the middle, the amount of image occlusion is increased, while the object occlusion is constant at 25\%. On the right, the amount of object occlusion is increased while the image occlusion is kept low. For the car class, a large improvement is observed over the entire occlusion spectrum. For the plane class the improvement is similarly significant, but absolute performance degenerates a bit sooner. This stems from the smaller object size and the related faster occlusion-to-object increase when increasing image occlusions. For the chair class, the improvement is less significant but the structural similarity is preserved for much longer. 
For the plane and car class the reconstruction quality is resilient to information loss (right) up to $\sim$50\%, where the decrease gains momentum. With the low amounts of image occlusion, the improvement is not significant for the chair class (consistent with left and middle).
}
\label{fig:recransacoccimg}
\end{figure}

%% file: figures/lfn_quali_combined.tex
\begin{figure}
\centering
\setlength\tabcolsep{6pt}
\begin{tabular}{>{\centering\arraybackslash} m{.45\textwidth} | >{\centering\arraybackslash} m{.45\textwidth}}
    \input{figures/lfn_quali_imgocc}

     & 
    \input{figures/lfn_quali}\\
\end{tabular}
    \caption{
    On the left, we show the qualitative effect of increasing occlusion on the same observation for the reconstruction of a novel view. Reconstructions of LFNs break early globally whereas RANRAC still provides a very decent reconstruction, only slowly introducing minor local (and natural/comprehensible) artifacts for completely hidden object parts. We further show the obtained consensus set, used for the final reconstruction (green inliers, red outliers).
    On the right, we show more qualitative results for novel view synthesis on different classes and the corresponding consensus sets.}
    \label{fig:lfn_quali_combined}
\end{figure}

%% file: figures/lfn_quali_imgocc.tex
\small
\includegraphics[width=\linewidth]{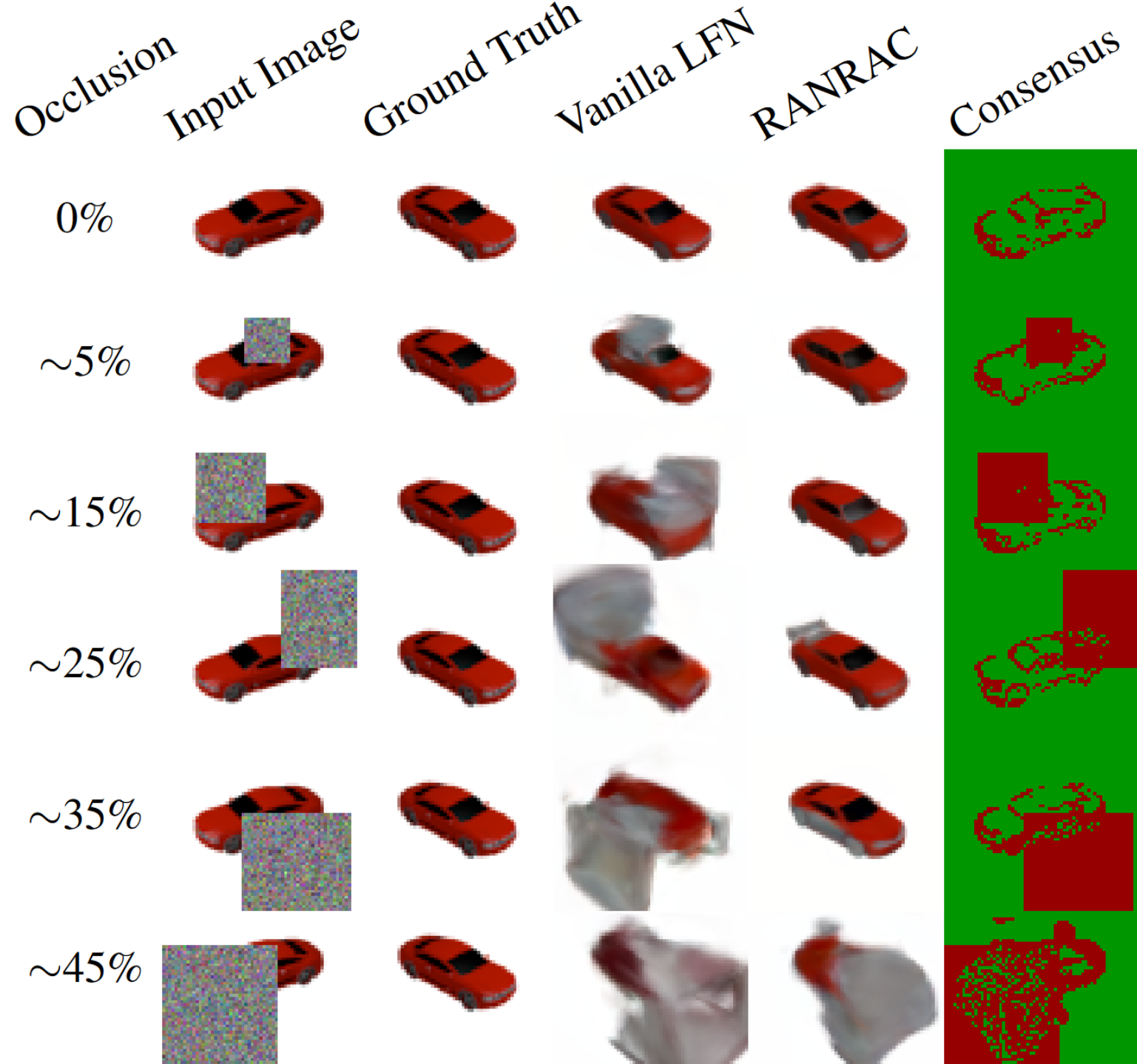}

%% file: figures/lfn_quali.tex
    \tiny
    \setlength\tabcolsep{0pt}
    \renewcommand{\arraystretch}{0}
    \newcommand{\qualiwidth}{\linewidth}
        \begin{tabular}{>{\centering\arraybackslash} m{1.1cm} >{\centering\arraybackslash} m{1.1cm} >{\centering\arraybackslash} m{1.1cm} >{\centering\arraybackslash} m{1.1cm} >{\centering\arraybackslash} m{1.1cm} }
            Input Image & G. Truth & Vanilla LFN & RANRAC & Consensus
        \\[.3cm]

            \includegraphics[width=\qualiwidth]{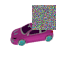}
            &
            \includegraphics[width=\qualiwidth]{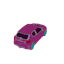}
            &
            \includegraphics[width=\qualiwidth]{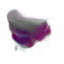}
            &
            \includegraphics[width=\qualiwidth]{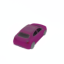}
            &
            \includegraphics[width=\qualiwidth]{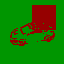}
        \\
            \includegraphics[width=\qualiwidth]{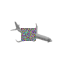}
            &
            \includegraphics[width=\qualiwidth]{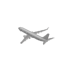}
            &
            \includegraphics[width=\qualiwidth]{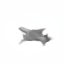}
            &
            \includegraphics[width=\qualiwidth]{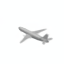}
            &
            \includegraphics[width=\qualiwidth]{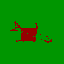}
        \\
            \includegraphics[width=\qualiwidth]{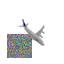}
            &
            \includegraphics[width=\qualiwidth]{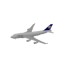}
            &
            \includegraphics[width=\qualiwidth]{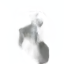}
            &
            \includegraphics[width=\qualiwidth]{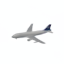}
            &
            \includegraphics[width=\qualiwidth]{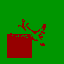}
        \\
            \includegraphics[width=\qualiwidth]{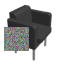}
            &
            \includegraphics[width=\qualiwidth]{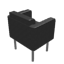}
            &
            \includegraphics[width=\qualiwidth]{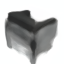}
            &
            \includegraphics[width=\qualiwidth]{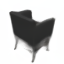}
            &
            \includegraphics[width=\qualiwidth]{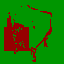}
        \\
        \end{tabular}
    \label{fig:results_quali}

%% file: figures/nerf_qual_new.tex
\begin{figure}
    \centering
    \includegraphics[width=0.95\linewidth]{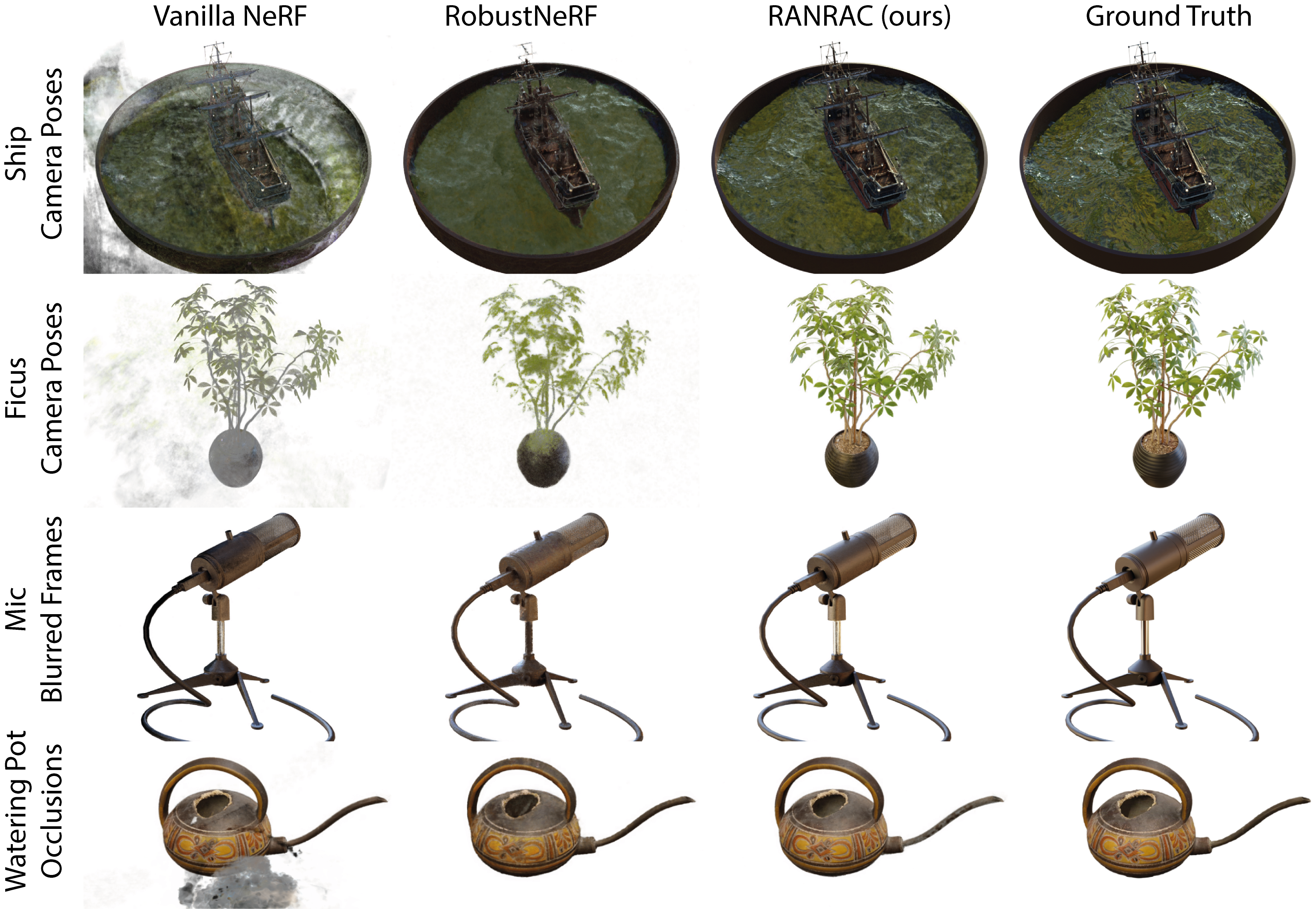}
    \caption{The occlusions lead to well-visible artifacts in the reconstructions of NeRF, these artifacts are completely removed by RANRAC. While RobustNeRF struggles with view-dependent and high-frequency details, RANRAC reliably reconstructs them.
    }
    \label{fig:nerf_floaters_removed}
\end{figure}

%% file: figures/nerf_table_combined.tex
\begin{table}
    \tiny
    \setlength\tabcolsep{1.5 pt}
    \centering

    \input{figures/nerf_table_joint}
    \vspace{.2cm}

    \input{figures/nerf_table_camera}
    \caption{
    RANRAC outperforms the state of the art RobustNeRF \cite{sabour2023robustnerf} and the baseline NeRF (without a method of robustness) for scenes contaminated with occlusions, blur and noisy camera parameters. We report PSNR$\uparrow$ averaged across perspectives and the $5$th percentile (Avg.|$P_5$) as artifacts introduced by inconsistencies only contaminate some views. We compare on the captured watering pot dataset with milder (10\%) and heavier (17.5\%) amounts of occluded perspectives and off-the-shelf datasets \cite{mildenhall2020nerf} with blurred perspectives and additive Gaussian noise $\mathcal{N}(5^{\circ}, 1^{\circ})$ on the camera parameters of 10\% of the perspectives. All three variants are built on top of instant-nsr for an isolated fair comparison of the robustness method. 
    Note that both robust approaches struggle separating the strong view-dependent effects of the ship scene from blur, leading to the exclusion of some perspectives (lower $P_5$ PSNR), while RANRAC still improves the overall reconstruction. For all other scenes and inconsistencies, RANRAC reliably separates inconsistent observations from clean ones. 
    }
    \label{tab:nerf_comined}
    \label{tab:nerf_eval}
    \label{tab:nerf_camera}
\end{table}

%% file: figures/nerf_table_joint.tex
    \begin{tabular}{c|cc|cccccccc}
         Inconsistency & Mild Occ. & Strong Occ. &
         \multicolumn{5}{c}{Blurred Perspectives}\\
          Dataset & \multicolumn{2}{c|}{Watering Pot} &
         Lego & Ship & Chair & Ficus & Mic\\
         \hline
         \textbf{RANRAC} &
         \bfseries 27.11|25.99 &
         \bfseries 26.12|24.94 &
         \bfseries 34.79|29.91 &
         \textbf{29.76}|16.14 &
         \bfseries 35.25|31.26 &
         \bfseries 31.36|28.54 &
         \bfseries 35.78|32.99\\
         
         
         RobustNeRF &
         26.83|25.58 &
         25.93|24.79 &
         29.14|23.69 &
         23.31|20.19 &
         31.11|27.12 &
         24.57|23.35 &
         30.21|26.85\\
         
         
         NeRF &
         26.65|22.61 &
         25.36|18.47 &
         31.15|19.00 &
         28.48|\textbf{20.35} &
         33.21|22.10 &
         29.04|20.12 &
         31.91|18.94\\
         
    \end{tabular}

%% file: figures/nerf_table_camera.tex
    \begin{tabular}{c|ccccccc}
         Inconsistency & \multicolumn{7}{c}{Noisy Camera Parameters}\\
         Dataset & Lego & Drums & Mic & Ship & Ficus & Hotdog & Materials\\
         \hline
         \textbf{RANRAC} &
         \bfseries 34.95|31.22 &
         \bfseries 25.88|22.77 &
         \bfseries 35.85|33.55 &
         \bfseries 30.77|22.05 &
         \bfseries 31.41|28.82 &
         \bfseries 37.16|30.53 &
         \bfseries 28.93|25.39\\
         
         RobustNeRF &
         29.83|26.82 &
         23.91|22.10 &
         29.36|27.55 &
         24.21|19.98 &
         23.17|21.85 &
         32.65|25.16 &
         24.90|22.30\\
         
         NeRF &
         24.67|15.95 &
         23.22|17.96 &
         28.82|22.15 &
         23.68|14.80 &
         28.12|22.77 &
         28.16|18.37 &
         24.73|19.89\\
    \end{tabular}

%% file: suple_aux/hp_additional_results.tex
\newlength{\widthmahp}
\setlength{\widthmahp}{0.5\textwidth}

\begin{figure}[!htb]
\centering
\begin{tikzpicture}
    \begin{axis}[
        width=\widthmahp,
        height=3.5cm,
        xticklabels={,,},
        ylabel={},
        legend pos=outer north east,
        legend style={/tikz/nodes={rotate=270}, font=\small, draw=none},
        name=planepsnr,
        xmajorgrids,
        ]
    
    \addplot+[mark options={scale=.5,solid},color=ibm2] coordinates {
        (30, 21.43)(40, 21.89)(50, 22.24)(60, 22.52)(70, 22.78)
        (80,22.98804965322117)(90,23.22999185803592)(100,23.390594480758196)
        };
    \addlegendentry{Plane}
    \addplot+[dashed,mark options={scale=.5,solid},color=ibm2] coordinates {
        (30, 24.86)(40, 25.11)(50, 25.29)(60, 25.63)(70, 25.66)
        (80, 25.8580922145922)(90,26.021008718474896)(100,26.168113444848775)
        };
    \addlegendentry{$P_{99}$}
    \legend{}
    \end{axis}
    \begin{axis}[
        width=\widthmahp,
        height=3.5cm,
        xticklabels={,,},
        ylabel={PSNR $\uparrow$},
        legend pos=outer north east,
        legend style={/tikz/nodes={rotate=270}, font=\small, draw=none},
        name=carpsnr,
        at=(planepsnr.south west), anchor=north west,
        xmajorgrids,
        ]

    \addplot+[mark options={scale=.5,solid},color=ibm3] coordinates {
        (30, 19.55)(40, 20.17)(50, 20.73)(60, 21.16)(70, 21.54)
        (80, 21.790497755930602)(90, 22.056263319081957)(100, 22.27532270425589)
        };
    \addlegendentry{Car}
    \addplot+[dashed, mark options={scale=.5,solid},color=ibm3] coordinates {
        (30, 23.54)(40, 23.70)(50, 24.14)(60, 24.45)(70, 24.71)
        (80, 24.84343821957416)(90, 24.97216164179912)(100, 25.13850163630571)
        };
    \addlegendentry{$P_{99}$}
    \legend{}
    
    \end{axis}
    \begin{axis}[
        width=\widthmahp,
        height=3.5cm,
        xticklabels={,,},
        ylabel={},
        legend pos=outer north east,
        legend style={/tikz/nodes={rotate=270}, font=\small, draw=none},
        name=chairpsnr,
        at=(carpsnr.south west), anchor=north west,
        xmajorgrids,
        ]

    \addplot+[mark options={scale=.5,solid},color=ibm4] coordinates {
        (30, 17.40)(40, 17.60)(50, 17.78)(60, 17.91)(70, 18.07)
        (80, 18.192960950783103)(90, 18.313104564515307)(100, 18.41428636435074)
        };
    \addlegendentry{Chair}
    \addplot+[dashed, mark options={scale=.5,solid},color=ibm4] coordinates {
        (30, 19.59)(40, 19.74)(50, 19.97)(60, 19.96)(70, 20.22)
        (80, 20.506563124223337)(90, 20.635312010479286)(100, 20.66378893329408)
        };
    \addlegendentry{$P_{99}$}
    \legend{}
    
    \end{axis}
    \begin{axis}[
        width=\widthmahp,
        height=3.5cm,
        xticklabels={,,},
        ylabel={},
        legend pos=outer north east,
        legend style={/tikz/nodes={rotate=270}, font=\small, draw=none},
        name=planessim,
        at=(chairpsnr.south west), anchor=north west,
        yshift=-0.25cm,
        xmajorgrids,
        ]
    
    \addplot+[mark options={scale=.5,solid}, color=ibm2] coordinates {
        (30, 0.7905)(40, 0.8033)(50, 0.8132)(60, .8210)(70, 0.8283)
        (80, 0.833425926264064)(90, 0.8398544110383879)(100, 0.8437911021203641)
        };
    \addlegendentry{Plane}
    \addplot+[dashed,mark options={scale=.5,solid}, color=ibm2] coordinates {
        (30, 0.8775)(40, 0.8826)(50, 0.8857)(60, .8904)(70, 0.8952)
        (80, 0.896858775031292)(90, 0.89856308109391)(100, 0.9003273064771626)
        };
    \addlegendentry{$P_{99}$}
    \legend{}
    
    \end{axis}
    \begin{axis}[
        width=\widthmahp,
        height=3.5cm,
        xticklabels={,,},
        ylabel={SSIM $\uparrow$},
        legend pos=outer north east,
        legend style={/tikz/nodes={rotate=270}, font=\small, draw=none},
        name=carssim,
        at=(planessim.south west), anchor=north west,
        xmajorgrids,
        ]

    \addplot+[mark options={scale=.5,solid}, color=ibm3] coordinates {
        (30, 0.8159)(40, 0.8309)(50, 0.8437)(60, 0.8525)(70, 0.8601)
        (80, 0.8656398367722942)(90, 0.8707223420909532)(100, 0.875003871961024)
        };
    \addlegendentry{Car}
    \addplot+[dashed, mark options={scale=.5,solid},color=ibm3] coordinates {
        (30, 0.8960)(40, 0.9014)(50, 0.9068)(60, 0.9093)(70, 0.9140)
        (80, 0.9160702203357289)(90, 0.9193130387925601)(100, 0.9190373698138088)
        };
    \addlegendentry{$P_{99}$}
    \legend{}
    \end{axis}
    \begin{axis}[
        width=\widthmahp,
        height=3.5cm,
        xlabel={\# samples},
        ylabel={},
        legend pos=outer north east,
        legend style={/tikz/nodes={rotate=270}, font=\small, draw=none},
        name=chairssim,
        at=(carssim.south west), anchor=north west,
        xmajorgrids,
        ]

    \addplot+[mark options={scale=.5,solid},color=ibm4] coordinates {
        (30, 0.6417)(40, 0.6504)(50, 0.6578)(60, 0.6625)(70, 0.6689)
        (80, 0.6738900972521776)(90, 0.6784242624423176)(100, 0.6821178568775382)
        };
    \addlegendentry{Chair}
    \addplot+[dashed, mark options={scale=.5,solid},color=ibm4] coordinates {
        (30, 0.7313)(40, 0.7369)(50, 0.7466)(60, 0.7460)(70, 0.7567)
        (80, 0.7649815393528013)(90, 0.772763637187511)(100, 0.7726331298547122)
        };
    \addlegendentry{$P_{99}$}
    
    \legend{}

    \end{axis}
\end{tikzpicture}
\hspace{0.5cm}
\begin{tikzpicture}
    \begin{axis}[
        width=\widthmahp,
        height=3.5cm,
        xticklabels={,,},
        ylabel={},
        legend pos=north east,
        legend style={font=\small, draw=none},
        name=planepsnr,
        xmajorgrids,
        ]
    
    \addplot+[mark options={scale=.5,solid},color=ibm2] coordinates {
(10, 24.84303688307026)(40, 25.369432158563463)(50, 25.507162216794104)(60, 25.369601644029068)(70, 25.44906147853721)(80, 25.182731094207504)(90, 25.46726381434021)(100, 25.506807758052123)(110, 25.43540601697737)(120, 25.49382207337965)(130, 25.3873400636984)(140, 25.303480122142876)(160, 24.87729175604104)(200, 24.55825334563236)(240, 24.029948963526664)(280, 23.54742098813226)
        };
    \addlegendentry{Plane}
    
    \end{axis}
    \begin{axis}[
        width=\widthmahp,
        height=3.5cm,
        xticklabels={,,},
        legend pos=south east,
        legend style={font=\small, draw=none},
        name=carpsnr,
        at=(planepsnr.south west), anchor=north west,
        xmajorgrids,
        ]

    \addplot+[mark options={scale=.5,solid},color=ibm3] coordinates {
(10, 23.298884829495996)(40, 24.030306668493452)(50, 23.745512203481198)(60, 24.236324006096808)(70, 24.108929392711733)(80, 24.289025749331298)(90, 24.448803936392714)(100, 24.242146341771786)(110, 24.16831695909096)(120, 24.243491417363156)(130, 24.290738772214123)(140, 24.109488164405224)(160, 24.53376937723213)(200, 24.302907406694715)(240, 24.110803781241994)(280, 24.245724196290986)
        };
    \addlegendentry{Car}
    
    \end{axis}
    \begin{axis}[
        width=\widthmahp,
        height=3.5cm,
        xticklabels={,,},
        ylabel={},
        legend pos=south east,
        legend style={font=\small, draw=none},
        name=chairpsnr,
        at=(carpsnr.south west), anchor=north west,
        xmajorgrids,
        ]

    \addplot+[mark options={scale=.5,solid},color=ibm4] coordinates {
(10, 17.840455611365208)(40, 18.645335842755003)(50, 18.91527501292694)(60, 19.053767807197154)(70, 19.08001712853323)(80, 19.169252243401857)(90, 19.197964123035234)(100, 19.247038247980978)(110, 19.49923006576365)(120, 19.5176222379714)(130, 19.377567431957118)(140, 19.31573694899084)(160, 19.538155158400897)(200, 19.665532384233643)(240, 19.75641641083956)(280, 19.70826034790494)
        };
    \addlegendentry{Chair}
    
    \end{axis}
    \begin{axis}[
        width=\widthmahp,
        height=3.5cm,
        xticklabels={,,},
        ylabel={},
        legend pos=north east,
        legend style={font=\small, draw=none},
        name=planessim,
        at=(chairpsnr.south west), anchor=north west,
        yshift=-0.25cm,
        xmajorgrids,
        scaled y ticks=false,
        yticklabel=\pgfkeys{/pgf/number format/.cd,fixed,precision=2,zerofill}\pgfmathprintnumber{\tick},
        ]
    
    \addplot+[mark options={scale=.5,solid}, color=ibm2] coordinates {
(10, 0.8913022975825966)(40, 0.8925535700412558)(50, 0.8966985044884538)(60, 0.8917525507155754)(70, 0.8934817626877609)(80, 0.8914937194511761)(90, 0.8958199052795469)(100, 0.8951309127218429)(110, 0.8959365071842433)(120, 0.8931341222288409)(130, 0.8927881339972367)(140, 0.8890016081554445)(160, 0.8833215195396618)(200, 0.8788824458259503)(240, 0.8743310966461191)(280, 0.864667943848527)
        };
    \addlegendentry{Plane}
    
    \end{axis}
    \begin{axis}[
        width=\widthmahp,
        height=3.5cm,
        xticklabels={,,},
        legend pos=south east,
        legend style={font=\small, draw=none},
        name=carssim,
        at=(planessim.south west), anchor=north west,
        xmajorgrids,
        scaled y ticks=false,
        yticklabel=\pgfkeys{/pgf/number format/.cd,fixed,precision=2,zerofill}\pgfmathprintnumber{\tick},
        ]

    \addplot+[mark options={scale=.5,solid}, color=ibm3] coordinates {
(10, 0.8942584749974752)(40, 0.9024478100254186)(50, 0.8985210891396931)(60, 0.9071777459093844)(70, 0.904695098981433)(80, 0.9059670241083746)(90, 0.9094782734787352)(100, 0.9078636816368402)(110, 0.9055546883745155)(120, 0.9055827818196113)(130, 0.9078723487208891)(140, 0.9036193843130094)(160, 0.9110978317664495)(200, 0.9069459570948051)(240, 0.9055457155263754)(280, 0.9079079348542513)
        };
    \addlegendentry{Car}

    \end{axis}
    \begin{axis}[
        width=\widthmahp,
        height=3.5cm,
        xlabel={\# samples},
        ylabel={},
        legend pos=south east,
        legend style={font=\small, draw=none},
        name=chairssim,
        at=(carssim.south west), anchor=north west,
        xmajorgrids,
        scaled y ticks=false,
        yticklabel=\pgfkeys{/pgf/number format/.cd,fixed,precision=2,zerofill}\pgfmathprintnumber{\tick},
        ]

    \addplot+[mark options={scale=.5,solid},color=ibm4] coordinates {
(10, 0.68806807373362)(40, 0.7172062198304802)(50, 0.7237199557090312)(60, 0.7303608118102529)(70, 0.7283294473163329)(80, 0.7309436986945567)(90, 0.7325340358219952)(100, 0.7356116987296838)(110, 0.7406980641746244)(120, 0.7425539575617057)(130, 0.7348421604617683)(140, 0.7409914441227181)(160, 0.7455702531836078)(200, 0.7494405559805821)(240, 0.7519026377228617)(280, 0.7520953945111619)
        };
    \addlegendentry{Chair}
    \end{axis}
\end{tikzpicture}
\caption{On the left, we show the reconstruction performance of LFNs on uniform-randomly down-sampled clean images. For every class and amount of samples, randomly selected instances have been inferred with 512 different draws of the samples each. Solid lines represent the average reconstruction performance, while dashed lines represent the average per-instance 99th percentile of the reconstruction performance, to emphasize the influence of the position of the samples. On the right, we show the reconstruction quality of RANRAC in the context of LFNs with varying amounts of samples.}
\label{fig:samplesparamransac}
\label{fig:reclfnsamples}
\end{figure}

\begin{figure}[!htb]
\centering
\begin{tikzpicture}
    \begin{axis}[
        width=\widthmahp,
        height=3.5cm,
        xticklabels={,,},
        ylabel={},
        legend pos=south west,
        legend style={font=\small, draw=none},
        name=planepsnr,
        xmajorgrids,
        ]
    
    \addplot+[mark options={scale=.5,solid},color=ibm2] coordinates {
(0.10, 24.891955997227726)(0.15, 25.208988225763335)(0.20, 25.193827122250287)(0.25, 25.369601644029068)(0.30, 25.022423951757162)(0.35, 24.8236768861595)(0.40, 23.256546064610585)
        };
    \addlegendentry{Plane}
    
    \end{axis}
    \begin{axis}[
        width=\widthmahp,
        height=3.5cm,
        xticklabels={,,},
        ylabel={PSNR $\uparrow$},
        legend pos=north east,
        legend style={font=\small, draw=none},
        name=carpsnr,
        at=(planepsnr.south west), anchor=north west,
        xmajorgrids,
        ]

    \addplot+[mark options={scale=.5,solid},color=ibm3] coordinates {
(0.10, 23.381622575792882)(0.15, 23.945946339858875)(0.20, 23.989948515240282)(0.25, 24.236324006096808)(0.30, 23.8880700325783)(0.35, 23.836189212927838)(0.40, 23.570928648914364)
        };
    \addlegendentry{Car}
    
    \end{axis}
    \begin{axis}[
        width=\widthmahp,
        height=3.5cm,
        xticklabels={,,},
        ylabel={},
        legend pos=north east,
        legend style={font=\small, draw=none},
        name=chairpsnr,
        at=(carpsnr.south west), anchor=north west,
        xmajorgrids,
        ]

    \addplot+[mark options={scale=.5,solid},color=ibm4] coordinates {
(0.10, 18.573964079070937)(0.15, 18.99008986035587)(0.20, 18.883156633737215)(0.25, 19.053767807197154)(0.30, 18.79379523743057)(0.35, 18.76205561034273)(0.40, 18.739683674932017)
        };
    \addlegendentry{Chair}
    
    \end{axis}
    \begin{axis}[
        width=\widthmahp,
        height=3.5cm,
        xticklabels={,,},
        ylabel={},
        legend pos=north east,
        legend style={font=\small, draw=none},
        name=planessim,
        at=(chairpsnr.south west), anchor=north west,
        yshift=-0.25cm,
        xmajorgrids,
        scaled y ticks=false,
        yticklabel=\pgfkeys{/pgf/number format/.cd,fixed,precision=3,zerofill}\pgfmathprintnumber{\tick},
        ]
    
    \addplot+[mark options={scale=.5,solid}, color=ibm2] coordinates {
(0.10, 0.8915431120288309)(0.15, 0.8965756660107536)(0.20, 0.8939406222025245)(0.25, 0.8917525507155754)(0.30, 0.8852229819109931)(0.35, 0.8805781990641515)(0.40, 0.8547050615714318)
        };
    \addlegendentry{Plane}
    
    \end{axis}
    \begin{axis}[
        width=\widthmahp,
        height=3.5cm,
        xticklabels={,,},
        ylabel={SSIM $\uparrow$},
        legend pos=north east,
        legend style={font=\small, draw=none},
        name=carssim,
        at=(planessim.south west), anchor=north west,
        xmajorgrids,
        scaled y ticks=false,
        yticklabel=\pgfkeys{/pgf/number format/.cd,fixed,precision=3,zerofill}\pgfmathprintnumber{\tick},
        ]

    \addplot+[mark options={scale=.5,solid}, color=ibm3] coordinates {
(0.10, 0.8948199921878491)(0.15, 0.902947914304438)(0.20, 0.9039254789154172)(0.25, 0.9071777459093844)(0.30, 0.9032299118426815)(0.35, 0.8999067522644811)(0.40, 0.898387391680293)
        };
    \addlegendentry{Car}

    \end{axis}
    \begin{axis}[
        width=\widthmahp,
        height=3.5cm,
        xlabel={Margin},
        ylabel={},
        legend pos=north east,
        legend style={font=\small, draw=none},
        name=chairssim,
        at=(carssim.south west), anchor=north west,
        xmajorgrids,
        scaled y ticks=false,
        yticklabel=\pgfkeys{/pgf/number format/.cd,fixed,precision=3,zerofill}\pgfmathprintnumber{\tick},
        ]

    \addplot+[mark options={scale=.5,solid},color=ibm4] coordinates {
(0.10, 0.7151994209380558)(0.15, 0.7253503178193109)(0.20, 0.7278854610220588)(0.25, 0.7303608118102529)(0.30, 0.7239825073345473)(0.35, 0.7192532941154542)(0.40, 0.7173006862372723)
        };
    \addlegendentry{Chair}
    \end{axis}
\end{tikzpicture}
%
\hspace{0.5cm}
\begin{tikzpicture}
    
    \begin{axis}[
        width=\widthmahp,
        height=3.5cm,
        xticklabels={,,},
        ylabel={},
        legend pos=south east,
        legend style={font=\small, draw=none},
        name=planepsnr,
        xmajorgrids,
        ]
    
    \addplot+[mark options={scale=.5,solid},color=ibm2] coordinates {
(512, 24.926881609156435)(1024, 25.032603400849617)(2048, 25.46726381434021)(4096, 25.346965244256154)
        };
    \addlegendentry{Plane}
    
    \end{axis}
    \begin{axis}[
        width=\widthmahp,
        height=3.5cm,
        xticklabels={,,},
        legend pos=south east,
        legend style={font=\small, draw=none},
        name=carpsnr,
        at=(planepsnr.south west), anchor=north west,
        xmajorgrids,
        ]

    \addplot+[mark options={scale=.5,solid},color=ibm3] coordinates {
(512, 24.054049274460123)(1024, 23.955142875999506)(2048, 24.448803936392714)(4096, 24.424545363078963)
        };
    \addlegendentry{Car}
    
    \end{axis}
    \begin{axis}[
        width=\widthmahp,
        height=3.5cm,
        xticklabels={,,},
        ylabel={},
        legend pos=south east,
        legend style={font=\small, draw=none},
        name=chairpsnr,
        at=(carpsnr.south west), anchor=north west,
        xmajorgrids,
        ]

    \addplot+[mark options={scale=.5,solid},color=ibm4] coordinates {
(512, 18.96574294150136)(1024, 19.115934411393116)(2048, 19.197964123035234)(4096, 19.216569939472468)
        };
    \addlegendentry{Chair}
    
    \end{axis}
    \begin{axis}[
        width=\widthmahp,
        height=3.5cm,
        xticklabels={,,},
        ylabel={},
        legend pos=south east,
        legend style={font=\small, draw=none},
        name=planessim,
        at=(chairpsnr.south west), anchor=north west,
        yshift=-0.25cm,
        xmajorgrids,
        scaled y ticks=false,
        yticklabel=\pgfkeys{/pgf/number format/.cd,fixed,precision=3,zerofill}\pgfmathprintnumber{\tick},
        enlarge y limits=0.2,
        ]
    
    \addplot+[mark options={scale=.5,solid}, color=ibm2] coordinates {
(512, 0.8858313879630216)(1024, 0.8876000091555266)(2048, 0.8958199052795469)(4096, 0.8923941143722184)
        };
    \addlegendentry{Plane}
    
    \end{axis}
    \begin{axis}[
        width=\widthmahp,
        height=3.5cm,
        xticklabels={,,},
        legend pos=south east,
        legend style={font=\small, draw=none},
        name=carssim,
        at=(planessim.south west), anchor=north west,
        xmajorgrids,
        scaled y ticks=false,
        yticklabel=\pgfkeys{/pgf/number format/.cd,fixed,precision=3,zerofill}\pgfmathprintnumber{\tick},
        enlarge y limits=0.2,
        ]

    \addplot+[mark options={scale=.5,solid}, color=ibm3] coordinates {
(512, 0.9050066211124786)(1024, 0.9032440795545614)(2048, 0.9094782734787352)(4096, 0.9089907831822693)
        };
    \addlegendentry{Car}

    \end{axis}
    \begin{axis}[
        width=\widthmahp,
        height=3.5cm,
        scaled y ticks=false,
        yticklabel=\pgfkeys{/pgf/number format/.cd,fixed,precision=3,zerofill}\pgfmathprintnumber{\tick},
        xlabel={Iterations},
        ylabel={},
        legend pos=south east,
        legend style={font=\small, draw=none},
        name=chairssim,
        at=(carssim.south west), anchor=north west,
        xmajorgrids,
        enlarge y limits=0.2,
        ]

    \addplot+[mark options={scale=.5,solid},color=ibm4] coordinates {
(512, 0.7238110073523775)(1024, 0.7281300956129515)(2048, 0.7325340358219952)(4096, 0.7313542021861669)
        };
    \addlegendentry{Chair}
    \end{axis}
\end{tikzpicture}
\caption{Reconstruction performance of RANRAC with different numbers of iterations (left) and different inlier margins (right).}
\label{fig:recransaciterations}
\label{fig:recransacmargin}
\end{figure}

\newlength{\widthmahpmargin}
\setlength{\widthmahpmargin}{0.12\textwidth}

\begin{figure}[!htb]
\small
\centering
\setlength\tabcolsep{1pt}
\renewcommand{\arraystretch}{0}
\begin{tabular}{ccccc}
    Input/Margin&
    0.1&
    0.25&
    0.35&
    0.4\\
    \includegraphics[width=\widthmahpmargin]{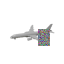}&
    \includegraphics[width=\widthmahpmargin]{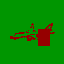}&
    \includegraphics[width=\widthmahpmargin]{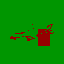}&
    \includegraphics[width=\widthmahpmargin]{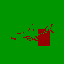}&
    \includegraphics[width=\widthmahpmargin]{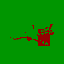}
    \\
    \includegraphics[width=\widthmahpmargin]{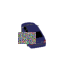}&
    \includegraphics[width=\widthmahpmargin]{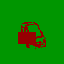}&
    \includegraphics[width=\widthmahpmargin]{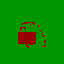}&
    \includegraphics[width=\widthmahpmargin]{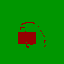}&
    \includegraphics[width=\widthmahpmargin]{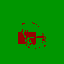}
    \\
    \includegraphics[width=\widthmahpmargin]{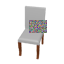}&
    \includegraphics[width=\widthmahpmargin]{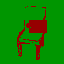}&
    \includegraphics[width=\widthmahpmargin]{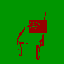}&
    \includegraphics[width=\widthmahpmargin]{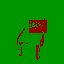}&
    \includegraphics[width=\widthmahpmargin]{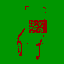}
    \\
\end{tabular}

\centering
\caption{Examples of the final consensus set produced by RANRAC for different margins (outliers are highlighted in red, inliers in green). Both effects of a larger margin are well visible: The increasing amount of occlusion that is not separated and the increasing amount of high-frequency details that are preserved. }
\label{fig:qualmargineffect}
\end{figure}

%% file: suple_aux/center_point_plot.tex
\pgfmathdeclarefunction{gauss}{3}{%
  \pgfmathparse{1/(#2*sqrt(2*pi))*exp(-((#3-#1)^2)/(2*#2^2))}%
}

\begin{figure}[!htb]
\begin{tikzpicture}
\begin{axis}[
    colormap/cool,
    xlabel={X},
    ylabel={Y},
    view={45}{70},
    width=0.4\textwidth,
]
\addplot3[
    mesh,
    samples=32,
    domain=0:63,
    y domain=0:63,
    point meta=z,
]
{
(0.5 * gauss(25,5,x) + 0.5 * gauss(39,5,x)) * (0.5 * gauss(25,5,y) + 0.5 * gauss(39,5,y))
};
\addlegendentry{Probability of Center Point}
\end{axis}
\end{tikzpicture}
\centering
\caption{Illustration of the probability of certain center points of the occlusion patch when drawn as described to achieve the desired partial occlusion of the objects.}
\label{fig:illustration_center_distribution}
\end{figure}